\documentclass[journal]{IEEEtran}
\usepackage{cite}

\usepackage{amsmath,amssymb,amsfonts,amsthm}
\usepackage{bm}
\usepackage{graphicx}
\usepackage{booktabs}
\usepackage{multirow}
\usepackage{algorithm}
\usepackage{algpseudocode}
\usepackage{xcolor}
\usepackage{cite}
\usepackage[normalem]{ulem}
\usepackage{url}

\usepackage{array}
\usepackage[caption=false]{subfig}

\usepackage{textcomp}
\usepackage{stfloats}
\usepackage{ragged2e}
\usepackage{verbatim}

\usepackage{hyperref}

\newif\ifinternal
\internalfalse

\newtheorem{theorem}{Theorem}
\newtheorem{assumption}{Assumption}
\newtheorem{remark}{Remark}

\newcommand{\R}{\mathbb{R}}
\newcommand{\E}{\mathbb{E}}
\newcommand{\Var}{\mathrm{Var}}
\newcommand{\tr}{\mathrm{tr}}
\newcommand{\diag}{\mathrm{diag}}
\newcommand{\KL}{\mathrm{KL}}
\newcommand{\Dx}{D_x}
\newcommand{\Dy}{D_y}
\newcommand{\bomega}{\bm{\omega}}
\newcommand{\bOmega}{\bm{\Omega}}
\newcommand{\On}{\bm{\Omega}_{\mathrm{n}}}
\newcommand{\Mn}{\mathbf{M}_{\mathrm{n}}}
\newcommand{\Mb}{\bar{\mathbf{M}}_{\mathrm{n}}}
\newcommand{\Sn}{\mathbf{S}_{\mathrm{n}}}
\newcommand{\En}{\mathbf{E}_{\mathrm{n}}}
\newcommand{\bL}{\mathbf{L}}
\newcommand{\bI}{\mathbf{I}}
\newcommand{\bK}{\mathbf{K}}
\newcommand{\bU}{\mathbf{U}}
\newcommand{\bx}{\mathbf{x}}
\newcommand{\by}{\mathbf{y}}
\newcommand{\bb}{\mathbf{b}}
\newcommand{\btheta}{\bm{\theta}}
\newcommand{\bTheta}{\bm{\Theta}}
\newcommand{\bphi}{\bm{\phi}}
\newcommand{\bSigma}{\bm{\Sigma}}
\newcommand{\mbar}{\bar{\mathbf{m}}}

\begin{document}

\title{Variational Orthogonal Random Features for Online Gaussian Processes}

\author{Yuhao~Liu,~\IEEEmembership{Member,~IEEE},
        Yuanqing~Song,~\IEEEmembership{Member,~IEEE},
        and~Petar~M.~Djuri\'{c},~\IEEEmembership{Life~Fellow,~IEEE}

\IEEEcompsocitemizethanks{
\IEEEcompsocthanksitem Y. Liu is an independent researcher, Fairfax, VA 22031 USA. Email: \href{mailto:yuhao1iu994@gmail.com}{yuhaoliu1994@gmail.com}. 

\IEEEcompsocthanksitem Y. Song is with the Department of Electrical \& Computer Engineering, Stony Brook University, Stony Brook, NY 11794 USA. 
E-mail: \href{mailto:yuanqing.song@stonybrook.edu}{yuanqing.song@stonybrook.edu}.

\IEEEcompsocthanksitem P. M. Djuri{\'c} is with the Department of Electrical \& Computer Engineering, Stony Brook University, Stony Brook, NY 11794 USA. 
E-mail: \href{mailto:petar.djuric@stonybrook.edu}{petar.djuric@stonybrook.edu}.
\par\vspace{1.2\baselineskip}
}
\thanks{This work was supported by the National Science Foundation (NSF) under Award 2212506.
}
}

\markboth{Submitted to  IEEE Transactions on Signal Processing}
{Y. Liu, Y. Song, and P. M. Djuri\'c}

\IEEEtitleabstractindextext{%
\begin{abstract}
Random Fourier features (RFFs) provide an efficient approximation of Gaussian processes (GPs) for large data sets and data streams. Standard RFFs draw the frequencies independently, and with a finite number of features these draws often cluster and leave parts of the spectrum uncovered. Orthogonal random features reduce this redundancy by hard constraints or by regularizers that lack a Bayesian interpretation. In this paper, we propose a hierarchical determinantal point process (H-DPP) prior that promotes spectral diversity through probabilistic repulsion and acts as a soft orthogonality prior on the frequency matrix. We show that the prior induces a Maxwell--Boltzmann spectral density, which corresponds to a band-pass Mexican hat kernel in the input domain. We also show that the widely used Frobenius-norm orthogonality penalty is a second-order approximation of the H-DPP evidence lower bound, which gives this penalty a probabilistic interpretation. We embed the prior in an online variational GP method that alternates recursive Bayesian updates of the weights with stochastic variational updates of the frequencies, and we prove that the recursive weight updates attain a cumulative regret of order $\log T$. On four UCI benchmarks, the method matches or improves the test RMSE of i.i.d.\ and hard-orthogonal random features, and on a data stream of 6.86 million flight records, it improves both RMSE and MNLL over online GP baselines.
\end{abstract}

}

\maketitle
\IEEEdisplaynontitleabstractindextext
\IEEEpeerreviewmaketitle

\section{Introduction}

\IEEEPARstart{G}{aussian} processes (GPs) are flexible nonparametric Bayesian models for regression and classification, and they quantify the uncertainty of their predictions explicitly. Their computational cost, however, grows cubically with the number of observations, which limits their use on large data sets and data streams. Two main approaches address this difficulty. Inducing-point methods~\cite{snelson2006sparse,titsias2009variational} build sparse function-space approximations from a small set of representative inputs, whereas random Fourier features (RFFs)~\cite{rahimi2007random} approximate stationary covariance functions by a finite-dimensional spectral representation. Because RFFs are simple, have a fixed per-step cost, and admit recursive Bayesian updates, they are particularly attractive for online GP inference~\cite{waxman2026sequential}. 

The quality of an RFF approximation depends on the distribution of the sampled frequencies. Standard RFFs draw the frequencies independently from the spectral density of the target kernel. Independent draws yield an unbiased Monte Carlo estimate of the kernel, but they do not control where the frequencies fall. Quasi-Monte Carlo (QMC) feature maps~\cite{avron2016quasi}
address the limitations of independent spectral samples
through low-discrepancy frequency constructions. With a finite number of features, several draws may land in nearby regions of the spectral domain while other regions receive little or no coverage. 
This redundancy lowers the information content of the features and raises the variance of the kernel approximation. It also limits the ability of the model to describe localized signal variations, such as transient events in sensor measurements and abrupt changes in nonstationary time series, which low-pass covariance functions tend to attenuate.

A natural remedy is to promote diversity among the spectral features. Orthogonal random features (ORFs)~\cite{yu2016orthogonal,choromanski2018geometry} replace independent frequencies with orthogonal directions. Specifically, ORFs enforce this diversity as a strict hard constraint over pre-set and non-learnable directions. Let $\bOmega=[\bomega_1,\ldots,\bomega_J]$ denote the matrix whose columns are the frequency vectors. Many machine learning methods encourage diversity among
learned representations through soft orthogonality
penalties, such as
\begin{equation*}
\|\bOmega^\top\bOmega-\bI\|_F^2,
\end{equation*}
where $\|\cdot\|_F$ denotes the Frobenius norm and
$\bOmega^\top\bOmega$ is the Gram matrix of the
column vectors of $\bOmega$. Other methods use
log-determinant regularizers to promote diversity
and linear independence among learned representations
\cite{guo2021orthogonal,xie2017learning}.

These penalties often improve robustness, representation quality, and stability of the optimization, with applications in graph neural networks~\cite{guo2021orthogonal}, neural signal analysis~\cite{angkan2025multi}, and fair representations~\cite{sarhan2020fairness}. Despite their empirical success, these penalties are largely heuristic. They specify a desired geometry of the feature space but do not explain why such a geometry should follow from a probabilistic model.

Determinantal point processes (DPPs)~\cite{kulesza2012determinantal} model
diversity through repulsive interactions among points. In GP models,
earlier work has used DPPs mainly to place inducing points in the input
space, with mixed results. On discrete domains, simulated annealing
outperforms DPP selection of inducing points~\cite{fortuin2021sparse}. A DPP
has also served as one of several priors on the inducing
inputs~\cite{rossi2021sparse}, and fixed-size $k$-DPP selection yields
provably accurate sparse variational
approximations~\cite{burt2020convergence}. In all these works, repulsion
acts on the inducing inputs, which summarize the posterior of a fixed
kernel. We instead let repulsion act in the spectral domain, where it
directly controls the distribution of the random features and thereby the
kernel representation itself.

In this paper, we develop a Bayesian model of spectral diversity with a hierarchical DPP (H-DPP) prior on the random frequencies. The prior encourages repulsion among spectral features, and its parameters can be optimized through variational inference. Unlike strict orthogonality constraints, the H-DPP prior defines a soft probability distribution over the frequency matrix, so the spectral representation can adapt to the data.

The proposed prior has two important consequences. First, we show that the spectral density that the H-DPP prior induces on each frequency is a Maxwell--Boltzmann density. The corresponding covariance function is the Mexican hat (Ricker) kernel, which has a band-pass response and emphasizes the localized signal variations that the radial basis function (RBF) kernel attenuates. Second, we prove that the Frobenius-norm orthogonality penalty is a second-order approximation of the H-DPP evidence lower bound (ELBO). Orthogonality regularization thus becomes a consequence of Bayesian inference rather than a heuristic choice.

We embed the prior in an online variational GP method. The method alternates recursive Bayesian updates of the regression weights with stochastic variational updates of the random frequencies. Its per-step cost is quadratic in the number of random features, and we prove a logarithmic bound on the cumulative regret of its recursive regression stage.

Our main contributions are as follows.
\begin{enumerate}
\item We introduce an H-DPP prior for RFFs that provides a Bayesian model of spectral diversity.
\item We show that the prior induces a Maxwell--Boltzmann spectral density and the corresponding Mexican hat kernel. These results connect spectral repulsion directly with band-pass covariance functions.
\item We give a probabilistic interpretation of widely used orthogonality regularizers. We show that the Frobenius-norm penalty is a second-order approximation of the H-DPP variational objective.
\item We propose an online variational GP method that adapts the spectral features and the regression weights jointly, prove a logarithmic regret bound for its recursive regression stage, and show that it matches or improves the RMSE of existing
random-feature methods on standard regression benchmarks and outperforms
online GP baselines on a large data stream.
\end{enumerate}

The rest of the paper is organized as follows. In Section~\ref{sec:related}, we review related work on random-feature and inducing-point GP approximations and on online GPs. In Section~\ref{sec:background}, we present background on GPs, Bochner's theorem, RFFs, and DPPs. In Section~\ref{sec:hdpp}, we introduce the H-DPP prior and derive its spectral properties. In Section~\ref{sec:vorf}, we develop our online variational inference method, and in Section~\ref{sec:analysis}, we analyze its regret and complexity. We report our experiments in Section~\ref{sec:experiments} and conclude in Section~\ref{sec:conclusions}.

\section{Related Work}\label{sec:related}

\subsection{RFF GPs}

RFFs approximate stationary kernels with frequencies drawn according to Bochner's theorem~\cite{rahimi2007random}. With $N$ observations and $J$ features, the cost of GP inference drops from $O(N^3)$ to $O(NJ^2)$, which is linear in the number of observations. Several works improve the choice of frequencies to reduce the variance of the kernel approximation. ORFs~\cite{yu2016orthogonal} replace independent frequencies with orthogonal directions. When the number of features exceeds the input dimension, ORFs concatenate several independently generated orthogonal blocks. Structured orthogonal random features (SORFs)~\cite{yu2016orthogonal} replace dense orthogonal matrices with products of structured transforms, such as Walsh--Hadamard matrices, and retain the accuracy of ORFs at a much lower cost. Theoretical results on the kernel approximation errors and spectral properties of orthogonal random features were established in~\cite{choromanski2018geometry}. 
Quasi-Monte Carlo (QMC) feature maps~\cite{avron2016quasi}
replace independent spectral samples with low-discrepancy
sequences to improve kernel approximation accuracy.
Unlike the proposed H-DPP approach, QMC methods do not
introduce a joint probabilistic repulsion prior over
the spectral frequencies.

Another line of work adapts the spectral representation
to the data. Sparse spectrum GPs~\cite{lazaro2010sparse}
optimize a finite set of frequencies, and variational
sparse spectrum GPs~\cite{gal2015improving} place
variational distributions over Fourier frequencies
and phases to account for spectral uncertainty.
Variational Fourier features~\cite{hensman2018variational}
instead use Fourier-domain inducing variables associated
with regularly spaced frequencies and perform variational
inference over the inducing variables rather than the
frequency locations. These methods do not promote
diversity among adaptive frequencies through a
joint probabilistic repulsion prior.

Our method combines adaptive spectral inference with
probabilistic repulsion among frequencies. Unlike methods
that impose hard orthogonality or use independent spectral
distributions, we place an H-DPP prior on the frequency
matrix. The resulting variational objective promotes
spectral diversity and allows the frequencies to adapt
to the data simultaneously.

\subsection{Inducing-Point GPs}

Inducing-point methods reduce the cost of GP inference with a small set of auxiliary variables that summarize the training data. Sparse variational GPs (SVGPs)~\cite{hensman2013gaussian,hensman2015scalable} cast the choice of inducing variables as variational inference and permit stochastic optimization with mini-batches. Later work increases the expressiveness of these approximations. Bayesian sparse GPs (BSGPs)~\cite{rossi2021sparse} account for uncertainty in the inducing variables, implicit posterior variational inference (IPVI)~\cite{yu2019implicit} uses implicit posteriors for more flexible variational inference, and stochastic gradient Hamiltonian Monte Carlo (SGHMC) draws samples from the posterior of deep GPs~\cite{havasi2018inference}.

These methods approximate the posterior through a sparse representation in the input space. Our method, in contrast, operates in the frequency domain. Rather than select inducing variables, it learns a spectral representation from the data under an H-DPP prior. It therefore modifies the kernel representation itself, whereas inducing-point methods approximate the posterior of a fixed kernel.

\subsection{Online GPs}

Online GP methods update the model as new observations arrive~\cite{waxman2026sequential}. Incremental sparse spectrum GPs (ISSGPs)~\cite{gijsberts2013real} perform
recursive updates of the weights with a fixed set of RFFs. Online sparse
GP regression (OSGPR)~\cite{bui2017streaming} propagates an approximate
posterior through streaming variational inference and also updates the
inducing inputs and the kernel hyperparameters as new data arrive.

ISSGPs keep the spectral representation fixed, so they cannot adapt it
when the frequency content of the data changes. OSGPR adapts the
hyperparameters of a fixed kernel family, such as its lengthscales, but not the shape of its spectral density. In addition, online GP methods commonly use the RBF kernel, whose Gaussian spectral density favors low frequencies and attenuates localized high-frequency variations.

Our method updates the regression weights and the spectral frequencies jointly by online variational inference. The Maxwell–Boltzmann marginal spectral density of the H-DPP prior corresponds to the Mexican hat kernel, whose band-pass response allows the model to represent the localized signal variations that RBF kernels suppress.

\section{Background}\label{sec:background}

\subsection{GPs and Bochner's Theorem}

A GP is a collection of random variables such that every finite subset has a joint Gaussian distribution. A GP $f(\bx)$ is specified by its mean function, which we assume to be zero throughout the paper, and by a positive-definite covariance function $\kappa(\bx,\bx')$.

We consider stationary covariance functions, for which the covariance depends only on the lag $\bm{\tau}=\bx-\bx'$. Bochner's theorem~\cite{rudin2017fourier} states that every continuous, stationary, positive-definite covariance function has the spectral representation
\begin{equation}\label{eq:bochner}
\kappa(\bm{\tau})=\int_{\R^{\Dx}} e^{i\bomega^\top\bm{\tau}}\,p(\bomega)\,d\bomega,
\end{equation}
where $p(\bomega)$ is a nonnegative spectral density that uniquely determines the covariance function. Stationary kernels and their spectral densities are therefore Fourier pairs.

The spectral density determines the smoothness and the frequency content of the covariance function. For example, the widely used RBF kernel has a Gaussian spectral density, which concentrates its mass near the origin and therefore favors low frequencies.

\subsection{Random Fourier Features}

To avoid the $O(N^3)$ cost of exact GPs, RFFs~\cite{rahimi2007random} approximate the integral in~\eqref{eq:bochner} by a Monte Carlo estimate. We draw $J$ frequencies $\{\bomega_j\}_{j=1}^J$ from the normalized spectral density $p(\bomega)$ and approximate the kernel by the inner product of a finite-dimensional feature map $\bphi(\bx)\in\R^J$,
\begin{equation}\label{eq:rff_kernel}
\kappa(\bx,\bx')\approx\bphi(\bx)^\top\bphi(\bx'),
\end{equation}
where
\begin{equation}\label{eq:rff_map}
\bphi(\bx)=\sqrt{\frac{2}{J}}\left[\cos(\bomega_1^\top\bx+b_1),\ldots,\cos(\bomega_J^\top\bx+b_J)\right]^\top,
\end{equation}
and the phases $b_j$ are drawn uniformly from $[0,2\pi]$. This form differs from the common construction with paired cosine and sine terms and no phase, but the two constructions have the same expectation. We explain our choice in Supplement~\ref{app:rff}. The feature map provides the GP approximation
\begin{equation}\label{eq:weight_space}
f(\bx)\approx\bphi(\bx)^\top\btheta,
\end{equation}
where $\btheta$ denotes the regression weights, whose prior distribution is specified in Section~5.1.

An important limitation of this approximation is that the frequencies are independent and identically distributed (i.i.d.). When $J$ is small, i.i.d.\ frequencies often cluster and cover the frequency domain poorly, which leaves gaps in the ability of the model to represent the underlying function.

\subsection{Determinantal Point Processes}

To address this problem, we turn to DPPs, which define distributions over sets of points that favor dissimilar, well-spread configurations. Formally, a simple point
process $W$ on a ground set
$\mathcal{G}\subseteq\R^{\Dx}$ is a DPP with correlation
kernel $k$ if its $J$-point correlation functions
satisfy~\cite{hough2006determinantal}
\begin{equation}
\rho_J(\bomega_1,\ldots,\bomega_J)
=\det(\bK_J),
\qquad
\bomega_1,\ldots,\bomega_J\in\mathcal{G},
\label{eq:dpp_corr}
\end{equation}
for every $J\ge 1$, where
$\bK_J\in\R^{J\times J}$ has entries
$[\bK_J]_{ij}=k(\bomega_i,\bomega_j)$.
The function $\rho_J$ describes the joint intensity
of $J$ distinct points with respect to the
Lebesgue measure.

In this paper, we consider configurations with a fixed
number $J$ of frequency vectors. We therefore use
a fixed-size L-ensemble, the continuous counterpart
of a $k$-DPP~\cite{kulesza2011kdpp}, in which
the density of a configuration
$\bOmega=[\bomega_1,\ldots,\bomega_J]$
is proportional to the determinant of a similarity
matrix. With the linear similarity
$\ell(\bomega_i,\bomega_j)
=\bomega_i^\top\bomega_j$, we have
\begin{equation}
p(\bOmega)\propto\det(\bOmega^\top\bOmega).
\label{eq:dpp_linear}
\end{equation}
This expression defines a normalizable density on a bounded ground set
when $J\le\Dx$. On the full space $\R^{\Dx\times J}$, it is not
integrable, and the prior in Section~\ref{sec:hdpp} adds a
Gaussian decay factor.

The main appeal of a DPP prior for RFFs is its repulsion. The determinant
$\det(\bOmega^\top\bOmega)$ vanishes whenever the frequencies are linearly
dependent, in particular when two of them coincide or point in the same or
opposite directions. The prior therefore pushes the frequency directions
apart and yields a more uniform coverage of the frequency domain.

\section{Hierarchical DPP Prior}\label{sec:hdpp}

We want to separate spectral diversity from the kernel lengthscales and, at the same time, retain compatibility with stochastic variational inference. To this end, we propose a hierarchical prior based on a decomposition into scale and direction,
\begin{equation}\label{eq:hier}
p(\bOmega,\bL)=p(\bOmega\mid\bL)\,p(\bL),
\end{equation}
where $\bL$ denotes the lengthscale matrix. When $\bL$ is diagonal, the model corresponds to an automatic relevance determination (ARD) kernel.

In standard RFF inference, the lengthscales and the frequencies are entangled, and their joint optimization is often unstable. To reduce these ambiguities, we introduce a latent spectral
configuration $\On\in\R^{\Dx\times J}$ in a standardized space,
where the subscript $n$ denotes normalization, and map it to
the physical frequencies by
\begin{equation}\label{eq:map}
\bOmega=\bL^{-1}\On.
\end{equation}
With this map, $\bL$ controls the coordinate-wise spectral scale,
whereas $\On$ controls the relative arrangement and diversity
of the frequencies.

A known drawback of the plain L-ensemble in~\eqref{eq:dpp_linear} is its bias toward vectors with large norms. Because $\det(\On^\top\On)$ grows with the volume spanned by the columns, the prior favors high frequencies. A constraint of the columns of $\On$ to the unit sphere $\mathcal{S}^{\Dx-1}$ would remove this bias, but such a hard constraint does not represent uncertainty in the norms. We therefore define a repulsive prior on all of $\R^{\Dx\times J}$ that balances a determinantal push against a Gaussian pull,
\begin{equation}\label{eq:hdpp}
p(\On)=\frac{1}{Z}\,\underbrace{\det(\On^\top\On)}_{\text{repulsive push}}\;\underbrace{\exp\!\left(-\frac{\Dx+1}{2}\|\On\|_F^2\right)}_{\text{Gaussian pull}},
\end{equation}
where $Z$ is the normalization constant. We give a procedure that draws samples from this prior in Supplement~\ref{app:sampling}. When $J>\Dx$, the Gram matrix $\On^\top\On$ is rank deficient and its determinant is zero. To keep the density well defined, we add a small jitter and use $\det(\On^\top\On+\epsilon\bI)$. In Supplement~\ref{app:overcomplete}, we provide more details.

The prior balances two competing effects. The determinant pushes the frequencies apart and prevents clusters, and the Gaussian term shrinks them toward the origin. The result is a stable distribution concentrated around the unit sphere. In Supplement~\ref{app:mode}, we prove that the norm of each column of $\On$ has its mode at one. This anchor lets $\On$ act as a descriptor of directions.

With the map in~\eqref{eq:map} and its Jacobian $|\det(\bL)|^J$, the density of the physical frequencies is
\begin{multline}\label{eq:hdpp_phys}
p(\bOmega\mid\bL)=\frac{|\det(\bL)|^J}{Z}\det(\bOmega^\top\bL^2\bOmega)\\
\times\exp\!\left(-\frac{\Dx+1}{2}\tr(\bOmega^\top\bL^2\bOmega)\right),
\end{multline}
where we assume that $\bL$ is symmetric. In practice, we place a noninformative (constant) prior on $\bL$.

\subsection{From Maxwellian Repulsion to Mexican Hat Kernels}
\label{ss:IV-A}

For $J\le\Dx$, we show that the marginal density of each
frequency $\bomega_j\in\R^{\Dx}$ under~\eqref{eq:hdpp_phys}
has the form
\begin{equation}\label{eq:maxwell}
p(\bomega_j)\propto
\|\bL\bomega_j\|_2^2
\exp\!\left(
-\frac{\Dx+1}{2}\|\bL\bomega_j\|_2^2
\right).
\end{equation}
This density is a dimension-normalized Maxwell density,
whose derivation is provided in Supplement~\ref{app:marginal}.
The supplement also discusses the case $J>\Dx$.

For the lag $\bm{\tau}=\bx-\bx'\in\R^{\Dx}$, the spectral
density in~\eqref{eq:maxwell} is a Gaussian density
multiplied by a quadratic function of the frequency.
By the Fourier-transform differentiation property,
its inverse Fourier transform, which defines the
covariance function, is proportional to the negative
Laplacian of a Gaussian in the transformed coordinates
$\bL^{-1}\bm{\tau}$. The resulting Mexican hat
covariance function, in normalized form, is
\begin{equation}\label{eq:mexhat}
\kappa_{\text{H-DPP}}(\bm{\tau})
=
\left(
1-\frac{\|\bL^{-1}\bm{\tau}\|_2^2}
{\Dx(\Dx+1)}
\right)
\exp\!\left(
-\frac{\|\bL^{-1}\bm{\tau}\|_2^2}
{2(\Dx+1)}
\right).
\end{equation}

Unlike the RBF kernel, whose spectral density is concentrated
near zero frequency, the H-DPP kernel has a band-pass
spectral response. Because its spectral density vanishes at
zero frequency, $p(\mathbf{0})=0$, the H-DPP kernel suppresses
low-frequency components. This property allows the model
to represent localized signal variations that may be
attenuated by the low-pass spectral response of the RBF kernel. 

Although standard RFFs can approximate the Mexican hat
kernel by drawing frequencies independently from its
spectral density, such sampling does not promote diversity
among the frequencies. In contrast, the H-DPP prior introduces 
repulsion through
the joint distribution of the frequencies and thereby
reduces spectral redundancy. For $J\le\Dx$, the marginal
spectral density remains unchanged.

\section{Online Variational Orthogonal Random Feature GPs}\label{sec:vorf}

We consider the model
\begin{align}
\by_t&=\bTheta_t^\top\bphi_t+\mathbf{e}_t,\label{eq:model}\\
\bphi_t&=\sqrt{\frac{2}{J}}\cos(\bOmega^\top\bx_t+\bb),\label{eq:feature}
\end{align}
where $\mathbf{e}_t\sim\mathcal{N}(\mathbf{0},\bSigma_e)$ and $\bSigma_e\in\R^{\Dy\times\Dy}$ is a diagonal covariance matrix. We assign an inverse-gamma prior to each diagonal element $\sigma_i^2$, so that the predictive distribution is a Student's $t$ distribution.

Our goal is to learn the weights $\bTheta_t=[\btheta_{i,t}]_{i=1}^{\Dy}\in\R^{J\times\Dy}$ and the spectral parameters $\{\bOmega,\bb,\bL\}$. We alternate between two stages. First, with the spectral parameters fixed, the model is linear in the weights, and we update the weights by exact Bayesian linear regression. Second, we refine the frequencies and the lengthscales by stochastic gradient descent (SGD) on the negative ELBO. In this way, the weights stay matched to the current features, while the features adapt to the signal dynamics under the H-DPP prior.

\subsection{Recursive Bayesian Regression in Feature Space}

For fixed spectral parameters $\{\bOmega,\bb,\bL\}$, the model reduces to a Bayesian linear model. To obtain closed-form Bayesian updates with unknown noise
variances, we adopt the conjugate normal--inverse-gamma
prior
\begin{equation}
\btheta_i\mid\sigma_i^2
\sim\mathcal{N}(\bm m_{i,0},\sigma_i^2\bSigma_0),
\qquad
\sigma_i^2\sim\mathcal{IG}(c_{i,0},d_{i,0}),
\end{equation}
for $i=1,\ldots,\Dy$, where $c_{i,0}$ and $d_{i,0}$
are the shape and scale parameters, respectively.
The posterior maintains the same form,
\begin{align}
\btheta_i\mid\sigma_i^2,\mathcal{D}_{1:t-1}
&\sim\mathcal{N}(\bm m_{i,t},\sigma_i^2\bSigma_t),\\
\sigma_i^2\mid\mathcal{D}_{1:t-1}
&\sim\mathcal{IG}(c_{i,t},d_{i,t}),
\end{align}
where $\mathcal{D}_{1:t-1}=\{(\bx_s,\by_s)\}_{s=1}^{t-1}$ denotes the
observations up to time $t-1$. 
After integration over the noise variance, the
predictive distribution is a Student's $t$
distribution.

Because all $\Dy$ outputs share the prior covariance $\bSigma_0$ and the
feature vector $\bphi_t$, they also share the matrix
$\bSigma_t\in\R^{J\times J}$, which reduces the cost of each update
considerably. We collect the remaining posterior parameters in
$\mathbf{M}_t=[\mathbf{m}_{i,t}]_{i=1}^{\Dy}$,
$\mathbf{c}_t=[c_{i,t}]_{i=1}^{\Dy}$, and
$\mathbf{d}_t=[d_{i,t}]_{i=1}^{\Dy}$, and we initialize
$\bSigma_1=\bSigma_0$ and these parameters with their prior values.
Algorithm~\ref{alg:rbr} summarizes the recursive update. It includes a
forgetting factor $\gamma\in(0,1]$ that discounts past observations in the
noise-variance statistics $\mathbf{c}_t$ and $\mathbf{d}_t$.

\begin{algorithm}[t]
\caption{Parallel Recursive Bayesian Update}\label{alg:rbr}
\begin{algorithmic}[1]
\Require Observation $(\bx_t,\by_t)$; spectral parameters $\{\bOmega,\bb,\bL\}$; prior state $\mathbf{M}_t,\bSigma_t,\mathbf{c}_t,\mathbf{d}_t$
\State \textbf{Feature map:} $\bphi_t=\sqrt{2/J}\,\cos(\bOmega^\top\bx_t+\bb)$
\State \textbf{Prediction:} $s_t=1+\bphi_t^\top\bSigma_t\bphi_t$
\Statex \hspace{\algorithmicindent}$\tilde{\by}_t=\by_t-\mathbf{M}_t^\top\bphi_t$
\State \textbf{Shared gain and covariance:} $\mathbf{k}_t=\bSigma_t\bphi_t/s_t$
\Statex \hspace{\algorithmicindent}$\bSigma_{t+1}=\bSigma_t-\mathbf{k}_t(\bSigma_t\bphi_t)^\top$
\State \textbf{Weights and noise:} $\mathbf{M}_{t+1}=\mathbf{M}_t+\mathbf{k}_t\tilde{\by}_t^\top$
\Statex \hspace{\algorithmicindent}$\mathbf{c}_{t+1}=\gamma\mathbf{c}_t+\tfrac{1}{2}\mathbf{1}$
\Statex \hspace{\algorithmicindent}$\mathbf{d}_{t+1}=\gamma\mathbf{d}_t+\tfrac{1}{2s_t}(\tilde{\by}_t\odot\tilde{\by}_t)$
\State \Return $\mathbf{M}_{t+1},\bSigma_{t+1},\mathbf{c}_{t+1},\mathbf{d}_{t+1}$
\end{algorithmic}
\end{algorithm}

\subsection{Variational Inference of RFFs with H-DPP Priors}

To optimize the spectral parameters, we infer the posterior
of $\{\On,\bL,\bb\}$. For simplicity, we take the posteriors
of the lengthscales $\bL$ and the phases $\bb$ to be
Dirac delta distributions. Their updates then reduce to
maximum a posteriori (MAP) steps by SGD, and the
variational inference concentrates on the standardized
frequency matrix $\On$.

We approximate the posterior of $\On$ by a factorized
Gaussian distribution $q(\On)$ and use the
reparameterization
\begin{equation}\label{eq:reparam}
\On=\Mb+\Sn\odot\En,\qquad
[\En]_{dj}\sim\mathcal{N}(0,1),
\end{equation}
where $\odot$ denotes the Hadamard (element-wise) product,
and $\Mb\in\R^{\Dx\times J}$ and
$\Sn\in\R^{\Dx\times J}$ are the variational mean and
standard deviation, respectively. 
 In practice, we keep the variational parameters
$\Mn=[\mathbf{m}_1,\ldots,\mathbf{m}_J]$ unconstrained
and project their columns onto the unit sphere,
$\mbar_j=\mathbf{m}_j/\|\mathbf{m}_j\|$. This projection separates the orientation of the spectrum, represented by $\On$, from the scale of the kernel, controlled by $\bL$. The matrix $\Sn$ represents the uncertainty in the spectral frequencies.
The Maxwell--Boltzmann density and the Mexican hat kernel in
Section~\ref{ss:IV-A} therefore describe the prior. Because
the projection fixes the norms of the variational means at one,
$q(\On)$ concentrates around the unit sphere when $\Sn$ is small, and the
spectrum of the learned features is set by the data through the
directions in $q(\On)$ and the scale matrix $\bL$. Variational sparse spectrum GPs~\cite{gal2015improving}
also account for uncertainty in the spectral frequencies.
Our approach differs through the joint H-DPP prior,
which introduces dependence among the frequencies
and promotes spectral diversity.

For RBF RFFs, a unit-norm constraint on the frequencies would
eliminate the radial variability of the Gaussian spectral
distribution and alter the low-pass response of the RBF kernel.
ORFs~\cite{yu2016orthogonal} avoid this problem: they orthogonalize
the frequency directions and rescale each direction by an
independent norm drawn from the $\chi$ distribution with $\Dx$
degrees of freedom. Each frequency therefore preserves its Gaussian
marginal distribution. However, the norms and directions in ORFs
are sampled once and remain fixed, and the directions satisfy
hard orthogonality constraints within each block.
For the unregularized H-DPP prior with $J\le\Dx$, the marginal
radial distribution has its mode at one
(Supplement~\ref{app:mode}). This property motivates the
unit-norm constraint on the variational means of the
standardized frequencies. The matrix $\bL$ controls the
coordinate-wise spectral scale and hence the input-domain
scale of the Mexican hat kernel induced by the prior.
In contrast to ORFs, the frequency directions can adapt to
the data under the soft orthogonality penalty derived from
the H-DPP prior.

\subsubsection{The ELBO}

We maximize the ELBO, which accounts for the predictive likelihood and for the structure of the H-DPP prior. 
We treat $\bL$ and $\bb$ as point estimates. With the noninformative prior
on $\bL$ and the uniform prior on $\bb$, their log-prior terms are
constant, so $\bL$ and $\bb$ enter the ELBO only through the likelihood
and act as hyperparameters. We express the ELBO by 
\begin{multline}\label{eq:elbo}
\mathcal{L}=\E_{q(\On)}\!\left[\ln p(\by\mid\On,\bL,\bb)\right]\\
-\KL\!\left(q(\On)\,\|\,p(\On)\right)+\text{const}.
\end{multline}
With the H-DPP prior in~\eqref{eq:hdpp}, the KL divergence expands as
\begin{multline}\label{eq:elbo2}
\mathcal{L}=\E_q[\ln p(\by\mid\On,\bL,\bb)]+\E_q[\ln\det(\On^\top\On+\epsilon\bI)]\\
-\frac{\Dx+1}{2}\E_q[\|\On\|_F^2]-\E_q[\ln q(\On)]+\text{const}.
\end{multline}

A major computational challenge with DPPs is the
$O(J^3)$ complexity of evaluating the log-determinant. Because the variational means lie on the unit sphere and the posterior uncertainties are usually small, we neglect terms of order $\|\Sn\|_F^4$. 
We then expand the log-determinant in~\eqref{eq:elbo2} to second order
about the identity matrix and take the expectation under
$q(\On)$ term by term. If we ignore the posterior uncertainty
($\Sn = \mathbf{0}$) and $\Mb$ is close to an orthonormal matrix,
the expansion reduces to
\begin{equation}\label{eq:taylor}
-\ln\det(\Mb^\top\Mb)\approx\frac{1}{2}\|\Mb^\top\Mb-\bI\|_F^2.
\end{equation}
This expression only illustrates the idea, because the full derivation must account for the expectation. We provide that derivation in Supplement~\ref{app:elbo}. The jitter $\epsilon$, commonly between $10^{-8}$ and $10^{-6}$, serves only numerical stability and has a negligible effect on the ELBO, so we let $\epsilon\to0$. Because the unit-norm constraint implies $\|\Mb\|_F^2=J$, we combine the preceding approximations and obtain 
\begin{multline}\label{eq:elbo_approx}
\mathcal{L}\approx\E_q[\ln p(\by\mid\On,\bL,\bb)]-\frac{1}{2}\|\Mb^\top\Mb-\bI\|_F^2\\
-\sum_{j,d}\left(\lambda S_{dj}^2-\ln S_{dj}\right),
\end{multline}
where $S_{dj}$ are the elements of $\Sn$ and $\lambda=\frac{\Dx-1}{2}+\frac{J+1}{\Dx}$. The detailed derivation is provided in Supplement D.

This result shows that the Frobenius-norm orthogonality penalty, common in point-estimation methods, is a second-order approximation of the repulsive DPP prior rather than a heuristic. With this soft orthogonality loss, we retain the diversity of the DPP at a cost of $O(J^2)$ instead of $O(J^3)$. In the SGD updates, we propagate the gradient with
respect to $\Mb$ back to the unconstrained parameters
$\Mn$ through the normalization map. This ensures
that the variational mean vectors remain on the unit
sphere after each update.

\begin{remark}\label{rem:overcomplete}
The expansion in~\eqref{eq:taylor} requires $\Mb^\top\Mb$ to be close to $\bI$, which is possible only when $J\le\Dx$. When $J>\Dx$, the Gram matrix has rank at most $\Dx$, and for unit-norm columns $\|\Mb^\top\Mb-\bI\|_F^2\ge J^2/\Dx-J$. In this regime, we preserve the penalty as a surrogate of the repulsive term. Its minimizers over unit-norm columns are the
unit-norm tight frames~\cite{benedetto2003finite},
which distribute the total projection energy
equally across all directions.
\end{remark}

\subsection{Predictive Inference and Uncertainty Decomposition}

For a new input $\bx_t$, we estimate the predictive
mean and variance of $\by_t$ by Monte Carlo averaging
over $q(\On)$, using samples
$\{\On^{(m)}\}_{m=1}^{N_{\mathrm{mc}}}$. By the law of total variance, the predictive variance of each output decomposes as
\begin{multline}\label{eq:totvar}
\sigma_{\mathrm{total}}^2=\Var_{q(\On)}\!\left(\E_{q(\btheta_t)}[y_t\mid\On]\right)\\
+\E_{q(\On)}\!\left(\Var_{q(\btheta_t)}[y_t\mid\On]\right).
\end{multline}
The first and second terms account for the uncertainty in the spectral features and in the weights and noise, respectively. At each time step, we predict $y_t$ before we use it to update the model, so that every prediction is a genuine out-of-sample forecast. Supplement~\ref{app:predictive} provides the details.

\section{Theoretical and Computational Analysis}\label{sec:analysis}

\subsection{Regret Analysis}

We now analyze the convergence of the recursive Bayesian regression in feature space. Consider a sequence of observations $\{(\bx_t,y_t)\}_{t=1}^T$ that arrive one at a time, and consider one output dimension. We consider the recursive Bayesian regression algorithm
with fixed spectral parameters and no forgetting
($\gamma=1$). We define its cumulative static regret
after $T$ steps as
\begin{align}
R_T&=\sum_{t=1}^T\left[-\ln p(y_t\mid y_{1:t-1},\bx_{1:t})\right]\nonumber\\
&\quad-\inf_{\btheta_r}\sum_{t=1}^T\left[-\ln p(y_t\mid\bx_t,\btheta_r)\right]\nonumber\\
&=-\ln p(y_{1:T}\mid\bx_{1:T})-\sum_{t=1}^T\left[-\ln p(y_t\mid\bx_t,\btheta_r^*)\right],\label{eq:regret}
\end{align}
where $\btheta_r=\{\btheta,\sigma^2\}$ denotes the regression parameters and $\btheta_r^*$ is the best static parameter in hindsight at time $T$. We make the following assumption.

\begin{assumption}\label{ass:ergodic}
The sequence $\{(\bx_t,y_t)\}$ is stationary and ergodic with
$\mathbb{E}[y_t^2]<\infty$. For the fixed spectral parameters, the
correlation matrix $\mathbf{R}_{\phi\phi}=\mathbb{E}[\bphi_t\bphi_t^\top]$ is
positive definite, and the minimum mean-square error
$\sigma_*^2=\min_{\btheta}\mathbb{E}[(y_t-\btheta^\top\bphi_t)^2]$ is
positive.
\end{assumption}

\begin{theorem}\label{thm:regret}
Let the spectral parameters $\{\On,\bL,\bb\}$ be fixed and let $\gamma=1$. Under Assumption~\ref{ass:ergodic}, the cumulative regret of the recursive Bayesian regression satisfies
\begin{equation}\label{eq:regret_bound}
R_T=\frac{J+1}{2}\ln T+O(1)\quad\text{almost surely}.
\end{equation}
\end{theorem}

We prove the theorem in Supplement~\ref{app:regret}. The coefficient $(J+1)/2$ equals one half of the number of free parameters, namely, $J$ weights and one noise variance. This agrees with the classical stochastic complexity of parametric models~\cite{rissanen1986stochastic}. The theorem applies only to the recursive regression stage
with fixed spectral features and does not extend to the
full algorithm with adaptive features.

\subsection{Computational Complexity}

We evaluate the cost of our variational orthogonal random feature (VORF) GP per time step. The total cost is
$O\big((N_{\mathrm{mc}}+\Dx)J^2 + N_{\mathrm{mc}}(\Dx+D_y)J\big)$,
where $N_{\mathrm{mc}}$ is the number of Monte Carlo samples.

\subsubsection{Prediction}
This stage computes the predictive mean and variance with Monte Carlo samples that account for the spectral uncertainty. The computational complexity of this stage is
\begin{equation}\label{eq:cost_pred}
T_{\mathrm{pred}}=O(N_{\mathrm{mc}}J^2+N_{\mathrm{mc}}\Dx J+N_{\mathrm{mc}}\Dy J).
\end{equation}
The quadratic term $O(N_{\mathrm{mc}}J^2)$ comes from the predictive variance $\bphi_t^\top\bSigma_t\bphi_t$ in Algorithm~\ref{alg:rbr} for each sample.

\subsubsection{Bayesian Update}
This stage performs the closed-form recursive update of the weight posterior and the noise hyperparameters. The recursive Bayesian update requires
\begin{equation}\label{eq:cost_bayes}
T_{\mathrm{Bayes}}=O(J^2+\Dy J),
\end{equation}
dominated by the rank-one update of the shared covariance matrix $\bSigma_t$.

\subsubsection{Spectral Update}
This stage updates the variational parameters $\{\Mn,\Sn,\bb,\bL\}$. Its cost is
\begin{equation}\label{eq:cost_opt}
T_{\mathrm{opt}}=O(\Dx J^2+\Dy J+\Dx J).
\end{equation}
The dominant term $O(\Dx J^2)$ comes from the Gram matrix $\Mb^\top\Mb$ in the gradient of the H-DPP term.

\section{Experiments}\label{sec:experiments}

In this section, we evaluate our VORF GP on real-world benchmarks. We pursue four goals.
\begin{itemize}
\item \emph{Value of the H-DPP prior.} We compare the i.i.d.\ RBF prior, the i.i.d.\ Mexican hat (Maxwell) prior, and our H-DPP prior, and we show that the H-DPP prior gives the most expressive features.
\item \emph{Soft versus hard orthogonality.}
We compare VORF with two hard-orthogonality methods:
orthogonal random features (ORF)~\cite{yu2016orthogonal}
and structured orthogonal random features
(SORF)~\cite{yu2016orthogonal}.
This comparison evaluates the benefit of adaptive
spectral features under a soft orthogonality penalty
relative to fixed orthogonal constructions.
\item \emph{Comparison with variational GPs.} Because we learn VORF by variational inference, we compare it with variational GPs based on inducing points. This comparison shows the efficiency of our weight-space approximation relative to function-space approximations.
\item \emph{Online performance.} We compare VORF with existing online GP methods, both inducing-point and RFF-based, on a large data stream.
\end{itemize}

For the first three studies, we use four regression data sets from the UCI Machine Learning Repository,\footnote{\url{https://archive.ics.uci.edu/}} namely, Concrete ($N=1{,}030$, $D=8$), Kin8nm ($N=8{,}192$, $D=8$), Powerplant ($N=9{,}568$, $D=4$), and Protein ($N=45{,}730$, $D=9$). For each data set, we shuffle the data with five random seeds and report the averages. All four data sets have scalar outputs, which suffice to demonstrate the main properties of our method. Our code is publicly available on GitHub.\footnote{\url{https://github.com/yuhaoliu94/VORF-GP}}

\subsection{Representational Power}

We first compare the i.i.d.\ RBF, i.i.d.\ Maxwell, and H-DPP priors on the Concrete data set with $J\in\{8,16,32,64,128\}$ random features.

\begin{figure*}[t]
\centering
\begin{minipage}{0.45\textwidth}\centering
\includegraphics[width=\textwidth]{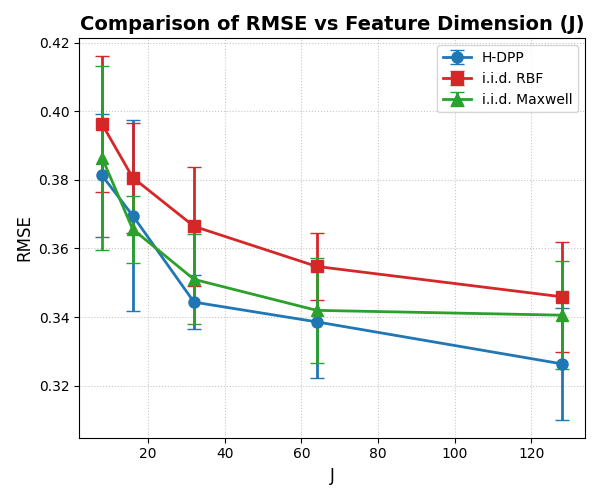}
\small (a) Normalized test RMSE versus $J$
\end{minipage}\hfill
\begin{minipage}{0.45\textwidth}\centering
\includegraphics[width=\textwidth]{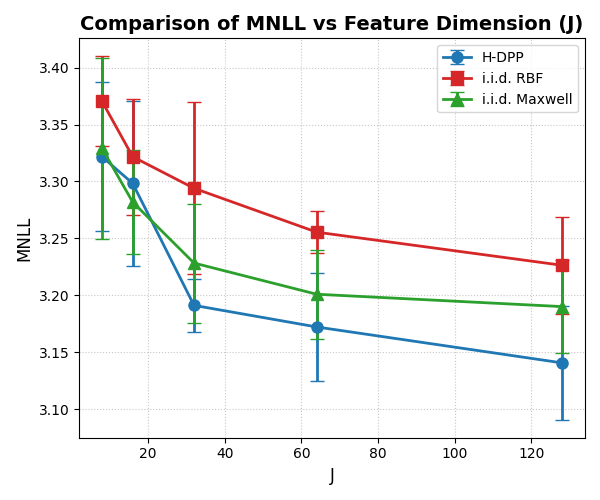}\\
\small (b) Test MNLL versus $J$
\end{minipage}
\caption{Performance of the i.i.d.\ RBF, i.i.d.\ Maxwell, and H-DPP priors on the Concrete data set.}
\label{fig:priors}
\end{figure*}

As Fig.~\ref{fig:priors} shows, the H-DPP prior attains the best RMSE and MNLL for almost all values of $J$. As $J$ grows, the performance of the i.i.d.\ Maxwell prior levels off, because its unconstrained features crowd together and become redundant. The H-DPP prior, in contrast, continues to improve with $J$, because its repulsion keeps the features apart and maintains their diversity.

To examine the features themselves, we study the empirical Gram matrix $\On^\top\On$ and its distance from the identity matrix. 
Table~\ref{tab:gram} quantifies this effect with the spectral approximation error $\|\On^\top\On-\bI\|_F$. The H-DPP prior has the smallest error for every $J$. These results show that the H-DPP prior produces more expressive features than its i.i.d.\ counterparts.

\begin{table}[t]
\centering
\caption{Spectral approximation error $\|\On^\top\On-\bI\|_F$ (std) for the three priors}
\label{tab:gram}
\small\setlength{\tabcolsep}{4pt}
\begin{tabular}{rccc}
\toprule
$J$ & i.i.d.\ RBF & i.i.d.\ Maxwell & H-DPP\\
\midrule
8   & 29.91 (7.17)       & 2.98 (0.43)  & \textbf{2.84} (0.13)\\
16  & 50.89 (1.99)       & 6.92 (0.54)  & \textbf{6.79} (1.02)\\
32  & 1448.95 (1942.95)  & 13.27 (1.25) & \textbf{12.02} (0.42)\\
64  & 309.77 (201.57)    & 25.22 (0.49) & \textbf{24.17} (0.40)\\
128 & 381.26 (2.63)      & 47.69 (0.61) & \textbf{45.90} (0.30)\\
\bottomrule
\end{tabular}
\end{table}

\subsection{Soft Versus Hard Orthogonality}

In this ablation study, we examine the effect of soft orthogonality. We compare VORF with ORF and SORF on the four UCI data sets. ORF and SORF use hard orthogonal random features, whereas VORF learns the spectral features under the soft orthogonality penalty that follows from the H-DPP prior. Table~\ref{tab:soft_hard} reports the normalized test RMSE. We do not report MNLL here, because ORF and SORF use a support vector machine (SVM) style predictor rather than a probabilistic GP likelihood, so their uncertainty estimates are not comparable with ours.

\begin{table}[t]
\centering
\caption{Normalized test RMSE (std) of soft and hard orthogonal features}
\label{tab:soft_hard}
\footnotesize\setlength{\tabcolsep}{3pt}
\begin{tabular}{lcccc}
\toprule
Method & Concrete & Kin8nm & Powerplant & Protein\\
\midrule
VORF & \textbf{0.33} (0.01) & \textbf{0.28} (0.00) & \textbf{0.23} (0.00) & \textbf{0.72} (0.00)\\
ORF  & 0.43 (0.04) & 0.61 (0.02) & \textbf{0.23 (0.00)} & 0.79 (0.01)\\
SORF & 0.45 (0.02) & 0.60 (0.02) & 0.24 (0.00) & 0.79 (0.00)\\
\bottomrule
\end{tabular}
\end{table}

As Table~\ref{tab:soft_hard} shows, VORF attains the lowest RMSE on three data
sets and ties with ORF on Powerplant. The gain is largest on Kin8nm, where VORF
reduces the RMSE by more than 50\% relative to both ORF and SORF. On Concrete,
VORF also improves clearly over both baselines. On Protein, the gain is smaller,
and on Powerplant the three methods perform almost equally. These results show that exact orthogonality can be too strict in practice. The soft penalty keeps the spectral features diverse and still lets them adapt to the data.

\subsection{Comparison with Variational GPs}

We next compare the VORF GP with several inducing-point variational GP methods, namely, BSGP, IPVI GP, SGHMC GP, and SVGP, on the same four data sets. We also include RFF GPs with i.i.d.\ RBF and i.i.d.\ Maxwell frequencies (RBFRF-GP and MaxwellRF-GP). Table~\ref{tab:mnll} summarizes the test MNLL and Table~\ref{tab:rmse} the normalized test RMSE. Lower values are better for both metrics. We take the baseline results for the inducing-point methods from the supplementary material\footnote{\url{https://proceedings.mlr.press/v130/rossi21a/rossi21a-supp.pdf}} of Rossi et al.~\cite{rossi2021sparse}.

\begin{table*}[t]
\centering
\begin{minipage}{0.48\textwidth}\centering
\caption{Test MNLL (std)}\label{tab:mnll}
\resizebox{\linewidth}{!}{%
\begin{tabular}{lcccc}
\toprule
Model & Concrete & Kin8nm & Powerplant & Protein\\
\midrule
BSGP$^\dagger$     & \textbf{3.12} (0.04) & \textbf{$-$1.12} (0.01) & \textbf{2.71} (0.02) & \textbf{2.78} (0.01)\\
IPVI GP$^\dagger$  & 3.19 (0.11) & \textbf{$-$1.12} (0.02) & 2.79 (0.03) & 2.81 (0.02)\\
SGHMC GP$^\dagger$ & 3.13 (0.09) & $-$1.08 (0.01) & 2.76 (0.05) & 2.81 (0.01)\\
SVGP$^\dagger$     & 3.18 (0.05) & $-$1.01 (0.01) & 2.79 (0.02) & 2.92 (0.01)\\
\midrule
RBFRF-GP     & 3.25 (0.07) & $-$1.10 (0.05) & 2.79 (0.02) & 2.93 (0.02)\\
MaxwellRF-GP & 3.21 (0.04) & $-$0.55 (0.44) & 2.81 (0.02) & 2.92 (0.01)\\
VORF-GP      & 3.14 (0.04) & $-$1.08 (0.05) & 2.81 (0.02) & 2.90 (0.01)\\
\bottomrule
\end{tabular}}
\end{minipage}\hfill
\begin{minipage}{0.48\textwidth}\centering
\caption{Normalized Test RMSE (std)}\label{tab:rmse}
\resizebox{\linewidth}{!}{%
\begin{tabular}{lcccc}
\toprule
Model & Concrete & Kin8nm & Powerplant & Protein\\
\midrule
BSGP$^\dagger$     & 0.40 (0.03) & 0.31 (0.01) & \textbf{0.23} (0.00) & \textbf{0.72} (0.00)\\
IPVI GP$^\dagger$  & 0.34 (0.02) & 0.31 (0.01) & \textbf{0.23} (0.00) & \textbf{0.72} (0.01)\\
SGHMC GP$^\dagger$ & 0.38 (0.03) & 0.34 (0.01) & \textbf{0.23} (0.00) & 0.75 (0.01)\\
SVGP$^\dagger$     & 0.35 (0.02) & 0.32 (0.00) & \textbf{0.23} (0.00) & 0.73 (0.00)\\
\midrule
RBFRF-GP     & 0.35 (0.02) & 0.30 (0.01) & \textbf{0.23} (0.00) & 0.74 (0.01)\\
MaxwellRF-GP & 0.35 (0.01) & 0.56 (0.21) & \textbf{0.23} (0.00) & 0.73 (0.01)\\
VORF-GP      & \textbf{0.33} (0.01) & \textbf{0.28} (0.00) & \textbf{0.23} (0.00) & \textbf{0.72} (0.00)\\
\bottomrule
\end{tabular}}
\end{minipage}\\[4pt]
\begin{minipage}{0.98\textwidth}\footnotesize
$^\dagger$Results taken from the supplementary material of~\cite{rossi2021sparse}.
\end{minipage}
\end{table*}

From Table~\ref{tab:mnll}, VORF-GP is competitive in terms of MNLL but
does not consistently outperform the inducing-point methods. In contrast, Table~\ref{tab:rmse} shows that VORF-GP achieves the lowest or matching RMSE across all four data sets.
The VORF GP attains the lowest RMSE on Concrete and Kin8nm and matches the best methods on Powerplant and Protein. These results suggest that the VORF GP reaches the accuracy of inducing-point variational GPs with a random-feature model that is easier to update online.

\subsection{Comparison with Online GPs}

Finally, we evaluate the VORF GP online on the Airline ArrDelay data set, which we build from the US Flights Data 2008\footnote{\url{https://www.kaggle.com/datasets/vikalpdongre/us-flights-data-2008}}. The target variable, ArrDelay, is the arrival delay in minutes, and the input has $\Dx=20$ variables. The cleaned data set contains about 6.86 million flight records, which makes it a large-scale test of online scalability. We split the data chronologically: the first 80\% of the records (5{,}484{,}023) form the training set and the last 20\% (1{,}371{,}006) the test set. All methods use 100 features or inducing points and update after each observation.

\begin{table}[t]
\centering
\caption{Online Regression Results on Airline ArrDelay (Lower Is Better)}
\label{tab:airline}
\begin{tabular}{lcc}
\toprule
Method & Normalized Test RMSE & Test MNLL\\
\midrule
ISSGP   & 0.6308 & 4.6684\\
OSGPR    & 0.4941 & 4.1224\\
VORF-GP & \textbf{0.2328} & \textbf{3.6835}\\
\bottomrule
\end{tabular}
\end{table}

Table~\ref{tab:airline} summarizes the results.
Relative to the best baseline, OSGPR, the VORF GP reduces the normalized
RMSE by about 53\% and the MNLL by 0.44 nats per test point. These
results show that our method performs well in large-scale online
regression, and that online adaptation of diverse spectral features
outperforms both a fixed sparse-spectrum representation (ISSGP) and a
streaming inducing-point approximation (OSGPR). Because the VORF GP
learns the spectral features together with the posterior, it requires
more wall-clock time than ISSGP and OSGPR.

\section{Conclusions}\label{sec:conclusions}

In this paper, we proposed a hierarchical DPP prior on the frequencies of random Fourier features. The prior separates the arrangement of the frequencies from the kernel scale and acts as a soft orthogonality prior. We showed that it induces a Maxwell--Boltzmann spectral density and hence a band-pass Mexican hat kernel, and that the Frobenius-norm orthogonality penalty is a second-order approximation of its ELBO. We embedded the prior in an online variational GP method that alternates exact recursive Bayesian updates of the weights with stochastic variational updates of the frequencies, and we proved that the recursive stage attains a cumulative regret of $\frac{J+1}{2}\ln T+O(1)$. In our experiments, the method matched or improved the test RMSE of i.i.d.\ and hard-orthogonal random features on four UCI benchmarks, matched or exceeded inducing-point variational GPs in RMSE, and outperformed online GP baselines on a data stream of 6.86 million records. In future work, we plan to extend the regret analysis to the full algorithm with adaptive features and to place informative priors on the scale matrix $\bL$. Because the columns of $\On$ have unit modal norm, $\bL$ also sets the frequency bands of the features, and a prior that restricts $\bL$ to a given band, such as the alpha band of the electroencephalogram, would let us decompose brain signals into physiologically meaningful components.

\bibliographystyle{IEEEtran}
\bibliography{ref.bib}

\clearpage
\newpage

\appendices

\section{The Form of the Random Features}\label{app:rff}
To approximate a stationary kernel $\kappa(\bx,\bx')$, we rely on Bochner's
theorem, which represents every continuous, positive-definite, stationary
kernel as the Fourier transform of a unique finite positive measure. We
assume that this measure has a density $p(\bomega)$ and that
$\kappa(\mathbf{0})=1$, so $p$ is a probability density. For a real-valued
kernel, this is equivalent to
\begin{align}
\kappa(\bx-\bx')&=\int_{\R^{\Dx}}p(\bomega)\cos\!\big(\bomega^\top(\bx-\bx')\big)\,d\bomega\nonumber\\
&=\E_{\bomega}\!\left[\cos\!\big(\bomega^\top(\bx-\bx')\big)\right].\label{eq:bochner_real}
\end{align}
We choose the feature map in~\eqref{eq:rff_map} so that it satisfies this expectation. Let $\hat\kappa(\bx,\bx')=\bphi(\bx)^\top\bphi(\bx')$ be the Monte Carlo estimate of $\kappa(\bx,\bx')$. For a single frequency ($J=1$), the identity $2\cos A\cos B=\cos(A-B)+\cos(A+B)$ gives
\begin{align}
\E_{\bomega,b}[\hat\kappa(\bx,\bx')]&=\E_{\bomega,b}\!\left[2\cos(\bomega^\top\bx+b)\cos(\bomega^\top\bx'+b)\right]\nonumber\\
&=\E_{\bomega}\!\left[\cos\!\big(\bomega^\top(\bx-\bx')\big)\right]\nonumber\\
&\quad+\E_{\bomega,b}\!\left[\cos\!\big(\bomega^\top(\bx+\bx')+2b\big)\right].\label{eq:unbiased}
\end{align}
Because $b$ is uniform on $[0,2\pi]$ with density $1/(2\pi)$, we have
\begin{equation}\label{eq:phase_zero}
\E_b\!\left[\cos(a+2b)\right]=\frac{1}{2\pi}\int_0^{2\pi}\cos(a+2b)\,db=0,
\end{equation}
where $a=\bomega^\top(\bx+\bx')$.
Thus, the scale $\sqrt{2}$ on each feature recovers the unit amplitude of $\kappa(\bx,\bx')$. Unlike the common construction with $J/2$ sine and $J/2$ cosine terms, our form attains the same unbiased estimate with $J$ cosine terms and evaluates only one type of basis function. 

For fixed frequencies and uniformly random phases, however, the paired
construction has a lower kernel-approximation variance for the Gaussian
kernel~\cite{sutherland2015error}. The reason is that each single-feature
estimate $2\cos(\bomega^\top\bx+b)\cos(\bomega^\top\bx'+b)$ contains the
term $\cos\big(\bomega^\top(\bx+\bx')+2b\big)$. By~\eqref{eq:phase_zero},
this term has zero mean over $b$, but it adds variance to the estimate.
In our model, the phases are not random draws. Their uniform distribution
is the prior, and it yields the stationary kernel in~\eqref{eq:mexhat}.
We learn the phases together with the frequencies, so the features no
longer act as a Monte Carlo estimate of $\kappa$, and the variance
comparison does not apply. Once the phases are learned, this term no longer averages out, and the
resulting covariance is nonstationary. In exchange, each feature can shift
its response, as we discuss next.

In the sine and cosine construction, the two weights of each pair set the
phase, separately for each output. In our construction, the phase $b_j$ is
a parameter of the feature itself and is shared by all outputs. With fixed
phases, a shift of the peak of a cosine would require a change of
$\bomega_j$, which stretches the whole function over the domain and can
impair the representation of long-range dependencies. A learned $b_j$
instead translates the response of a feature without a change of its
period, which allows precise models of time lags and spatial offsets.

For numerical stability, we wrap each phase after
every gradient step,
\begin{equation}\label{eq:mod}
b_j\leftarrow b_j\pmod{2\pi},
\end{equation}
which restricts the phases to $[0,2\pi)$ without
changing the feature values. For $J$ frequencies,
the cosine representation requires $JD_y$ regression
weights and $J$ phase parameters, compared with
$2JD_y$ regression weights in the paired sine
and cosine representation. Thus, the total number
of parameters is unchanged for scalar outputs
and decreases by $J(D_y-1)$ for multiple outputs.

\section{Samples from the H-DPP Prior}\label{app:sampling}

In this Supplement, we describe how we draw samples $\On\sim p(\On)$.

\subsection{Radial Distribution}

The prior is
\begin{equation*}
p(\On)\propto\det(\On^\top\On)\exp\!\left(-\frac{\Dx+1}{2}\tr(\On^\top\On)\right).
\end{equation*}
We change to hyperspherical coordinates, and for each column, we write $\bomega_j=r_j\mathbf{u}_j$ with $r_j=\|\bomega_j\|_2$ and $\mathbf{u}_j\in\mathcal{S}^{\Dx-1}$, so that the volume element in $\R^{\Dx}$ becomes
\begin{equation}\label{eq:volume}
d\bomega_j=r_j^{\Dx-1}\,dr_j\,d\sigma(\mathbf{u}_j).
\end{equation}
With $\bU=[\mathbf{u}_1,\ldots,\mathbf{u}_J]\in\R^{\Dx\times J}$, the Gram determinant factorizes as $\det(\On^\top\On)=\big(\prod_{j=1}^Jr_j^2\big)\det(\bU^\top\bU)$, and so does the joint density,
\begin{equation}\label{eq:factor}
p\big(\{r_j,\mathbf{u}_j\}_{j=1}^J\big)\propto\underbrace{\prod_{j=1}^J r_j^{\Dx+1}e^{-\frac{\Dx+1}{2}r_j^2}}_{\text{radial part}}\;\underbrace{\det(\bU^\top\bU)}_{\text{angular part}}.
\end{equation}
The radial density of each $r_j$ is therefore $p(r_j)\propto r_j^{\Dx+1}\exp\!\big(-\frac{\Dx+1}{2}r_j^2\big)$. The chi distribution with $k$ degrees of freedom has density proportional to $r^{k-1}e^{-r^2/2}$, so the H-DPP prior induces a scaled chi distribution. Specifically, $\sqrt{\Dx+1}\,r_j\sim\chi_{\Dx+2}$.

\subsection{Procedure}

For $J\le\Dx$, the factorization in~\eqref{eq:factor}
provides an exact generative procedure in three steps.

\begin{enumerate}
\item \emph{Directions (angular DPP layer).}
We generate $J$ unit vectors
$\{\mathbf{u}_j\}_{j=1}^J$ sequentially.
The first direction $\mathbf{u}_1$ is drawn uniformly
from $\mathcal{S}^{\Dx-1}$.
For $k=2,\ldots,J$, we draw a candidate
$\mathbf{v}$ uniformly from the unit sphere
and accept it with probability
\begin{equation}
a_k(\mathbf{v})
=1-\|\bm Q_{k-1}^\top\mathbf{v}\|_2^2,
\end{equation}
where $\bm Q_{k-1}$ contains an orthonormal
basis for the span of the previously accepted
directions. Rejected candidates are replaced
by new draws.

\item \emph{Norms (Maxwell layer).}
Independently draw
$\hat r_j\sim\chi_{\Dx+2}$ and set
$r_j=\hat r_j/\sqrt{\Dx+1}$.

\item \emph{Assembly.}
Form
$\On=[r_1\mathbf{u}_1,\ldots,r_J\mathbf{u}_J]$.
\end{enumerate}

The sequential procedure samples exactly from the
angular determinantal distribution with density
proportional to $\det(\bU^\top\bU)$, because the
Gram determinant factorizes into the product of
the squared orthogonal residuals.

For $J>\Dx$, the Gram matrix $\bU^\top\bU$
is rank deficient, and its determinant vanishes.
The radial and angular components no longer
factorize under the jitter-regularized H-DPP prior.
We therefore sample them jointly using the
Metropolis-within-Gibbs procedure described
in Supplement~B.3. In practice, we also use
this regularized procedure for $J\le\Dx$
to improve numerical stability.

\subsection{\texorpdfstring{Regularization for Overcomplete Sets ($J>\Dx$)}{Regularization for Overcomplete Sets}}\label{app:overcomplete}

To keep the measure well defined when $J>\Dx$, we add a small constant $\epsilon$ inside the determinant and use $\det(\On^\top\On+\epsilon\bI_J)$. We write $\On=\bU\mathbf{R}$, where $\bU=[\mathbf{u}_1,\ldots,\mathbf{u}_J]$ collects the directions on $\mathcal{S}^{\Dx-1}$ and $\mathbf{R}=\diag(r_1,\ldots,r_J)$ the norms. The target density is
\begin{equation}\label{eq:reg_target}
p(\mathbf{R},\bU)\propto\det(\mathbf{R}\bU^\top\bU\mathbf{R}+\epsilon\bI_J)\prod_{j=1}^J r_j^{\Dx-1}e^{-\frac{\Dx+1}{2}r_j^2}.
\end{equation}
This density does not factorize, so we draw samples from it with a Metropolis-within-Gibbs scheme. For each index $j\in\{1,\ldots,J\}$ in turn, we perturb the norm and the direction of the $j$th column jointly.
\begin{enumerate}
\item \emph{Norm.} We propose $r_j^*=r_j\exp(\sigma_r\zeta)$ with $\zeta\sim\mathcal{N}(0,1)$.
\item \emph{Direction.} We propose $\mathbf{u}_j^*=\mathrm{Proj}_{\mathcal{S}^{\Dx-1}}(\mathbf{u}_j+\sigma_u\bm{\eta})$ with $\bm{\eta}\sim\mathcal{N}(\mathbf{0},\bI_{\Dx})$.
\item \emph{Acceptance.} We accept $(\mathbf{u}_j^*,r_j^*)$ with probability $\alpha=\min\!\big(1,\exp(\Delta\mathcal{L}+\ln(r_j^*/r_j))\big)$, where
\begin{multline}\label{eq:mh}
\Delta\mathcal{L}=\ln\frac{\det(\mathbf{R}^*\bU^{*\top}\bU^*\mathbf{R}^*+\epsilon\bI_J)}{\det(\mathbf{R}\bU^\top\bU\mathbf{R}+\epsilon\bI_J)}\\
+(\Dx-1)\ln\frac{r_j^*}{r_j}-\frac{\Dx+1}{2}\big((r_j^*)^2-r_j^2\big)
\end{multline}
is the change of the log target, and $\ln(r_j^*/r_j)$ is the Hastings correction for the log-normal proposal. The projected Gaussian proposal on the sphere is symmetric and needs no correction.
\end{enumerate}

In practice, we use the jitter-regularized prior
for sample generation, including when $J\le\Dx$,
to improve numerical stability. For $J\le\Dx$,
we subsequently take $\epsilon\to0$ in the
analytical derivations. When $J>\Dx$, the
unregularized Gram determinant vanishes, and
we use the soft-orthogonality surrogate discussed
in Remark~1 for variational optimization.

\section{Properties of the Marginal Spectral Density}\label{app:marginal}

Recall that $\On=[\bomega_{1},\ldots,\bomega_{J}]\in\R^{\Dx\times J}$, where we drop the subscript of the columns for brevity, and
\begin{equation}\label{eq:prior_again}
p(\On)=\frac{1}{Z}\det(\On^\top\On)\exp\!\left(-\frac{\Dx+1}{2}\|\On\|_F^2\right).
\end{equation}
The joint density is invariant under permutations of the columns, so all columns have the same marginal density. We now derive this marginal.

\emph{Case $J\le\Dx$.} By~\eqref{eq:factor}, the norms $r_1,\ldots,r_J$ are mutually independent and independent of the directions $\bU$, and each has density proportional to $r^{\Dx+1}e^{-\frac{\Dx+1}{2}r^2}$. The angular part $\det(\bU^\top\bU)$ is invariant under a common rotation $\bU\mapsto\mathbf{Q}\bU$ for every orthogonal matrix $\mathbf{Q}$, because $(\mathbf{Q}\bU)^\top(\mathbf{Q}\bU)=\bU^\top\bU$. Hence the marginal distribution of each direction $\mathbf{u}_j$ is rotation invariant, that is, uniform on $\mathcal{S}^{\Dx-1}$. With the volume element~\eqref{eq:volume}, the density
of $\bomega_j$ with respect to the Lebesgue measure is
therefore
\begin{align}
p(\bomega_j)
&\propto
\frac{r_j^{\Dx+1}
e^{-\frac{\Dx+1}{2}r_j^2}}
{r_j^{\Dx-1}}\nonumber\\
&=\|\bomega_j\|_2^2
\exp\!\left(
-\frac{\Dx+1}{2}\|\bomega_j\|_2^2
\right),\label{eq:marg_exact}
\end{align}

\emph{Case $J>\Dx$.} With the jitter, the angular argument still holds, because $\det(\mathbf{R}\bU^\top\mathbf{Q}^\top\mathbf{Q}\bU\mathbf{R}+\epsilon\bI)$ is also invariant under common rotations. The marginal of each column therefore remains isotropic. The norms, however, no longer separate from the directions, so the radial profile of the marginal differs from that in~\eqref{eq:marg_exact}. In this regime, we regard~\eqref{eq:maxwell} as the design density of each column rather than as its exact marginal.

\subsection{Mode of the Marginal Norm}\label{app:mode}

Let $r_j=\|\bomega_j\|_2$. By~\eqref{eq:marg_exact}
and the volume element~\eqref{eq:volume}, the radial
density is
\begin{equation}\label{eq:radial}
f(r_j)\propto r_j^{\Dx-1}p(\bomega_j)
=r_j^{\Dx+1}
\exp\!\left(-\frac{\Dx+1}{2}r_j^2\right).
\end{equation}
The mode $r_j^*$ of the norm maximizes $f(r_j)$.
The derivative of the logarithm is
\begin{align}
\frac{d}{dr_j}\ln f(r_j)
&=\frac{d}{dr_j}
\left[(\Dx+1)\ln r_j
-\frac{\Dx+1}{2}r_j^2\right]\nonumber\\
&=\frac{\Dx+1}{r_j}-(\Dx+1)r_j.
\label{eq:dlogf}
\end{align}
At the critical point,
\begin{equation}\label{eq:mode}
\frac{\Dx+1}{r_j^*}-(\Dx+1)r_j^*=0
\quad\Longrightarrow\quad
(r_j^*)^2=1.
\end{equation}
Because the norm is nonnegative, the mode of the
standardized norm is $r_j^*=1$.

\subsection{Mode Under the Regularized H-DPP}

We examine how the jitter regularization affects
the mode of the radial distribution. For the
single-frequency case $J=1$, we show that a
small regularization parameter $\epsilon$ changes
the mode only slightly from its unregularized
value of one.

Consider a single column $\bomega$ with $J=1$. With the jitter, the density becomes
\begin{equation}\label{eq:reg_single}
p(\bomega)\propto(\|\bomega\|^2+\epsilon)\exp\!\left(-\frac{\Dx+1}{2}\|\bomega\|^2\right).
\end{equation}
With $r=\|\bomega\|$ and the volume element~\eqref{eq:volume}, the radial density is
\begin{equation}\label{eq:reg_radial}
f(r)\propto\big(r^{\Dx+1}+\epsilon r^{\Dx-1}\big)\exp\!\left(-\frac{\Dx+1}{2}r^2\right).
\end{equation}
We set the derivative of $\ln f(r)$ to zero,
\begin{equation}\label{eq:reg_deriv}
\frac{(\Dx+1)r^{\Dx}+\epsilon(\Dx-1)r^{\Dx-2}}{r^{\Dx+1}+\epsilon r^{\Dx-1}}-(\Dx+1)r=0,
\end{equation}
which simplifies to
\begin{equation}\label{eq:reg_simpl}
(\Dx+1)r^2+\epsilon(\Dx-1)=(\Dx+1)r^2(r^2+\epsilon).
\end{equation}
With $u=r^2$, we obtain the quadratic equation
\begin{equation}\label{eq:quad}
(\Dx+1)u^2-(\Dx+1)(1-\epsilon)u-\epsilon(\Dx-1)=0,
\end{equation}
whose positive root is
\begin{equation}\label{eq:reg_mode}
(r^*)^2=\frac{(1-\epsilon)+\sqrt{(1-\epsilon)^2+\frac{4\epsilon(\Dx-1)}{\Dx+1}}}{2}.
\end{equation}
For $\epsilon=0$, the equation reduces to $(\Dx+1)u^2-(\Dx+1)u=0$, so $u=1$ and $r^*=1$. A first-order expansion with $\sqrt{1+x}=1+\frac{x}{2}+o(x)$ gives
\begin{equation}\label{eq:reg_mode_approx}
r^*=1-\frac{\epsilon}{\Dx+1}+o(\epsilon).
\end{equation}
With $\epsilon$ between $10^{-8}$ and $10^{-6}$, as in our experiments, the deviation from one is negligible.

Thus, for $J=1$, the jitter regularization has
a negligible effect on the mode of the radial
distribution when $\epsilon$ is small. This result
provides additional motivation for constraining
the variational means of the standardized
frequencies to the unit sphere.

\section{Derivation of the ELBO}\label{app:elbo}
In this Supplement, we derive the ELBO of our model and show how the
repulsive H-DPP prior leads to the soft orthogonality penalty. With point
estimates of $\bL$ and $\bb$, the objective is
\begin{multline}\label{eq:elbo_full}
\mathcal{L}=\E_{q(\On)}\big[\ln p(\by\mid\On,\bL,\bb)\big]
-\mathrm{KL}\big(q(\On)\,\|\,p(\On)\big)\\
+\ln p(\bL)+\ln p(\bb),
\end{multline}
and its maximization over $\bL$ and $\bb$ corresponds to MAP estimation.
The prior on $\bL$ is noninformative and the prior on $\bb$ is uniform,
so the last two terms are constants. We obtain
\begin{align}
\mathcal{L}&=\underbrace{\E_q[\ln p(\by\mid\On,\bL,\bb)]}_{\text{Term 1}}\underbrace{-\E_q[\ln q(\On)]}_{\text{Term 2}}\nonumber\\
&\quad+\underbrace{\E_q[\ln\det(\On^\top\On+\epsilon\bI)]}_{\text{Term 3}}\nonumber\\
&\quad\underbrace{-\frac{\Dx+1}{2}\E_q[\tr(\On^\top\On)]}_{\text{Term 4}}+\text{const}.\label{eq:elbo_terms}
\end{align}
Under the variational posterior~\eqref{eq:reparam}, the elements $\omega_{dj}\sim\mathcal{N}(\bar M_{dj},S_{dj}^2)$ are mutually independent. We also use the following notation:
\begin{itemize}
\item The mean vectors lie on the unit sphere, $\|\mbar_j\|_2^2=\sum_{d=1}^{\Dx}\bar M_{dj}^2=1$ for all $j$.
\item The Gram matrix is $\bK=\On^\top\On$, and its expectation is $\bar\bK=\E_q[\bK]$.
\item The total variance of the $j$th frequency vector is $v_j=\|\mathbf{s}_j\|_2^2=\sum_{d=1}^{\Dx}S_{dj}^2$.
\end{itemize}

\subsection{Trace Term}

With $\E[X^2]=\mu^2+\sigma^2$, we have
\begin{align}
\text{Term 4}&=-\frac{\Dx+1}{2}\sum_{j=1}^J\sum_{d=1}^{\Dx}\E_q[\omega_{dj}^2]\nonumber\\
&=-\frac{\Dx+1}{2}\sum_{j=1}^J\Big(\underbrace{\textstyle\sum_d\bar M_{dj}^2}_{1}+\underbrace{\textstyle\sum_dS_{dj}^2}_{v_j}\Big)\nonumber\\
&=-\frac{\Dx+1}{2}J-\frac{\Dx+1}{2}\sum_{j=1}^Jv_j.\label{eq:term4}
\end{align}

\subsection{Entropy Term}

For the factorized Gaussian $q(\On)=\prod_{d,j}\mathcal{N}(\bar M_{dj},S_{dj}^2)$, the entropy is
\begin{equation}\label{eq:term2}
\text{Term 2}=\sum_{j=1}^J\sum_{d=1}^{\Dx}\frac{1}{2}\ln(2\pi eS_{dj}^2)=\sum_{j=1}^J\sum_{d=1}^{\Dx}\ln S_{dj}+\text{const}.
\end{equation}

\subsection{Repulsion Term}

We need $\text{Term 3}=\E_q[\ln\det(\bK+\epsilon\bI)]$. When we perform a second-order expansion about $(1+\epsilon)\bI$, we obtain
\begin{multline}\label{eq:logdet_exp}
\ln\det(\bK+\epsilon\bI)=J\ln(1+\epsilon)+\frac{1}{1+\epsilon}\tr(\bK-\bI)\\
-\frac{1}{2(1+\epsilon)^2}\|\bK-\bI\|_F^2+o(\|\bK-\bI\|_F^2).
\end{multline}
The term $J\ln(1+\epsilon)$ is a constant that vanishes as $\epsilon\to0$. For the remaining terms, we need the moments of $\bK$. Because the elements are independent, $\E[\omega_{di}\omega_{dj}]=\E[\omega_{di}]\E[\omega_{dj}]$ for $i\neq j$, and the off-diagonal elements of $\bar\bK$ are
\begin{equation}\label{eq:Koff}
\bar K_{ij}=\sum_{d=1}^{\Dx}\bar M_{di}\bar M_{dj}=\mbar_i^\top\mbar_j.
\end{equation}
For the diagonal elements, the unit-norm constraint implies
\begin{equation}\label{eq:Kdiag}
\bar K_{jj}=\sum_{d=1}^{\Dx}(\bar M_{dj}^2+S_{dj}^2)=1+v_j.
\end{equation}
Thus,
\begin{equation}\label{eq:Kbar}
\bar\bK=\Mb^\top\Mb+\diag(v_1,\ldots,v_J),
\end{equation}
and the expected trace is
\begin{equation}\label{eq:trace}
\E_q[\tr(\bK-\bI)]=\tr(\bar\bK-\bI)=\sum_{j=1}^Jv_j.
\end{equation}

\subsubsection{Expected Frobenius Norm}

Using the identity $\E[X^2]=\mu^2+\sigma^2$,
we obtain the decomposition
\begin{equation}\label{eq:frob_decomp}
\E_q[\|\bK-\bI\|_F^2]=\|\bar\bK-\bI\|_F^2+\sum_{i,j}\Var_q(K_{ij}).
\end{equation}
The diagonal of $\Mb^\top\Mb-\bI$ is zero, so by~\eqref{eq:Kbar}
\begin{equation}\label{eq:frob_mean}
\|\bar\bK-\bI\|_F^2=\|\Mb^\top\Mb-\bI\|_F^2+\sum_{j=1}^Jv_j^2.
\end{equation}
For the off-diagonal variances, independence across
$d$ implies that
$\Var_q(K_{ij})=\sum_d\Var_q(\omega_{di}\omega_{dj})$. With $\Var(XY)=\E[X^2]\E[Y^2]-(\E[X])^2(\E[Y])^2$ for independent $X$ and $Y$, we obtain
\begin{equation}\label{eq:var_off}
\Var_q(K_{ij})=\sum_{d=1}^{\Dx}\left(\bar M_{di}^2S_{dj}^2+\bar M_{dj}^2S_{di}^2+S_{di}^2S_{dj}^2\right),\quad i\neq j.
\end{equation}
For the diagonal elements $K_{jj}=\sum_d\omega_{dj}^2$, we use $\Var(X^2)=4\mu^2\sigma^2+2\sigma^4$ for $X\sim\mathcal{N}(\mu,\sigma^2)$, which follows from $\E[X^4]=\mu^4+6\mu^2\sigma^2+3\sigma^4$ and $(\E[X^2])^2=(\mu^2+\sigma^2)^2$. Hence
\begin{equation}\label{eq:var_diag}
\Var_q(K_{jj})=\sum_{d=1}^{\Dx}\left(4\bar M_{dj}^2S_{dj}^2+2S_{dj}^4\right).
\end{equation}
Altogether,
\begin{multline}\label{eq:var_total}
\sum_{i,j}\Var_q(K_{ij})=\sum_{i\neq j}\sum_{d=1}^{\Dx}\left(\bar M_{di}^2S_{dj}^2+\bar M_{dj}^2S_{di}^2+S_{di}^2S_{dj}^2\right)\\
+\sum_{j=1}^J\sum_{d=1}^{\Dx}\left(4\bar M_{dj}^2S_{dj}^2+2S_{dj}^4\right),
\end{multline}
and the expansion of Term 3 reads
\begin{multline}\label{eq:term3_exact}
\E_q[\ln\det(\bK+\epsilon\bI)]=J\ln(1+\epsilon)+\frac{1}{1+\epsilon}\sum_{j=1}^Jv_j\\
-\frac{1}{2(1+\epsilon)^2}\Big(\|\Mb^\top\Mb-\bI\|_F^2+\sum_{j=1}^Jv_j^2+\sum_{i,j}\Var_q(K_{ij})\Big)\\
+\E_q\big[o(\|\bK-\bI\|_F^2)\big].
\end{multline}

The exact variance in~\eqref{eq:var_total} couples $\bar M_{dj}^2$ and $S_{dj}^2$. This form still permits gradients at a cost of $O(\Dx J)$, but we simplify it further to separate the two sets of parameters. We adopt a decoupled approximation in the spirit of mean-field methods and assume that the squared means and the variances are empirically uncorrelated across the dimensions $d$,
\begin{align}
\sum_{d=1}^{\Dx}\bar M_{di}^2S_{dj}^2&=\Big(\sum_{d=1}^{\Dx}\bar M_{di}^2\Big)\Big(\frac{1}{\Dx}\sum_{d=1}^{\Dx}S_{dj}^2\Big)+\Dx\,\widehat{\mathrm{Cov}}(\mbar_i^2,\mathbf{s}_j^2)\nonumber\\
&=\frac{v_j}{\Dx}+\Dx\,\widehat{\mathrm{Cov}}(\mbar_i^2,\mathbf{s}_j^2)\approx\frac{v_j}{\Dx},\label{eq:decouple}
\end{align}
where $\widehat{\mathrm{Cov}}$ denotes the sample covariance over $d$ and we use the unit-norm constraint. With this approximation and without the terms of order $S^4$, the total variance simplifies to
\begin{align}
\sum_{i,j}\Var_q(K_{ij})&\approx\sum_{i\neq j}\left(\frac{v_j}{\Dx}+\frac{v_i}{\Dx}\right)+\sum_{j=1}^J\frac{4v_j}{\Dx}\nonumber\\
&=\frac{2(J-1)}{\Dx}\sum_{j=1}^Jv_j+\frac{4}{\Dx}\sum_{j=1}^Jv_j\nonumber\\
&=\frac{2(J+1)}{\Dx}\sum_{j=1}^Jv_j.\label{eq:var_simpl}
\end{align}
We substitute this result into~\eqref{eq:term3_exact}, drop $v_j^2$ as a term of order $S^4$, and let $\epsilon\to0$. The expected log-determinant becomes
\begin{align}
\E_q[\ln\det(\bK)]&\approx\sum_{j=1}^Jv_j-\frac{1}{2}\|\Mb^\top\Mb-\bI\|_F^2
-\frac{J+1}{\Dx}\sum_{j=1}^Jv_j\nonumber\\
&=\Big(1-\frac{J+1}{\Dx}\Big)\sum_{j=1}^Jv_j-\frac{1}{2}\|\Mb^\top\Mb-\bI\|_F^2.\label{eq:term3_final}
\end{align}
The repulsion, thus, acts on the variational uncertainty with a strength set by the ratio of the number of frequencies $J$ to the input dimension $\Dx$.

\subsection{Final Objective}

We now combine the four terms of~\eqref{eq:elbo_terms}. With~\eqref{eq:term4}, \eqref{eq:term2}, and~\eqref{eq:term3_final}, we obtain
\begin{align}
\mathcal{L}&\approx\E_q[\ln p(\by\mid\On,\bL,\bb)]+\sum_{j,d}\ln S_{dj}\nonumber\\
&\quad+\Big(1-\frac{J+1}{\Dx}\Big)\sum_{j=1}^Jv_j-\frac{1}{2}\|\Mb^\top\Mb-\bI\|_F^2\nonumber\\
&\quad-\frac{\Dx+1}{2}\sum_{j=1}^Jv_j+\text{const}\nonumber\\
&=\E_q[\ln p(\by\mid\On,\bL,\bb)]+\sum_{j,d}\left(\ln S_{dj}-\lambda S_{dj}^2\right)\nonumber\\
&\quad-\frac{1}{2}\|\Mb^\top\Mb-\bI\|_F^2+\text{const},\label{eq:elbo_final}
\end{align}
where $\lambda=\frac{\Dx-1}{2}+\frac{J+1}{\Dx}$. We minimize the negative ELBO,
\begin{multline}\label{eq:loss}
\ell_{\mathrm{total}}=\underbrace{-\E_q[\ln p(\by\mid\On,\bL,\bb)]}_{\text{NLL}}+\underbrace{\frac{1}{2}\|\Mb^\top\Mb-\bI\|_F^2}_{\text{orthogonality penalty}}\\
+\underbrace{\sum_{j,d}\left(\lambda S_{dj}^2-\ln S_{dj}\right)}_{\text{variance regularizer}}.
\end{multline}
Without data, the variance regularizer drives each $S_{dj}$ toward the
common value $1/\sqrt{2\lambda}$. The likelihood term then reduces all
$S_{dj}$ further. In both cases, the variances vary little across $d$,
and their sample covariance with the squared means is small.

\section{Monte Carlo Predictive Inference}\label{app:predictive}

To account for the uncertainty in the random features, we use Monte Carlo integration for prediction. Given the current posterior of the weights $p(\bTheta\mid\mathcal{D}_{1:t-1})$ and the variational distribution $q(\On)$, the predictive distribution of a new observation $\by_t\in\R^{\Dy}$ is
\begin{multline}\label{eq:pred_int}
p(\by_t\mid\bx_t,\mathcal{D}_{1:t-1})=\iint p(\by_t\mid\bx_t,\bTheta,\On)\\
\times p(\bTheta\mid\mathcal{D}_{1:t-1})\,q(\On)\,d\bTheta\,d\On.
\end{multline}
We approximate this integral with samples $\{\On^{(m)}\}_{m=1}^{N_{\mathrm{mc}}}$ from $q(\On)$. For each sample, the feature vector is $\bphi_t^{(m)}=\sqrt{2/J}\,\cos(\bOmega^{(m)\top}\bx_t+\bb)$, with $\bOmega^{(m)}=\bL^{-1}\On^{(m)}$. The predictive mean is the average over the samples,
\begin{equation}\label{eq:pred_mean}
\hat\by_t\approx\frac{1}{N_{\mathrm{mc}}}\sum_{m=1}^{N_{\mathrm{mc}}}\E_{\bTheta}[\by_t\mid\bx_t,\bTheta,\On^{(m)}]=\frac{1}{N_{\mathrm{mc}}}\sum_{m=1}^{N_{\mathrm{mc}}}\mathbf{M}_t^\top\bphi_t^{(m)}.
\end{equation}
By the law of total variance,
\begin{align}
\Var(\by_t\mid\bx_t)&=\Var_{\On}\!\big(\E_{\bTheta}[\by_t\mid\On]\big)+\E_{\On}\!\big(\Var_{\bTheta}[\by_t\mid\On]\big)\nonumber\\
&=\Var_{\On}\!\big(\mathbf{M}_t^\top\bphi_t\big)+\E_{\On}\!\big(\Var_{\bTheta}[\by_t\mid\On]\big).\label{eq:totvar_app}
\end{align}
For output $i$, the variance in the second term is
\begin{align}
\Var(y_{i,t}\mid\bphi_t)&=\E_{\btheta_{i,t},\sigma_{i,t}^2}\!\big[\Var(y_{i,t}\mid\bphi_t,\btheta_{i,t},\sigma_{i,t}^2)\big]\nonumber\\
&\quad+\Var_{\btheta_{i,t},\sigma_{i,t}^2}\!\big(\E[y_{i,t}\mid\bphi_t,\btheta_{i,t},\sigma_{i,t}^2]\big)\nonumber\\
&=\E[\sigma_{i,t}^2]+\Var_{\btheta_{i,t},\sigma_{i,t}^2}\!\big(\bphi_t^\top\btheta_{i,t}\big)\nonumber\\
&=\E[\sigma_{i,t}^2]+\E_{\sigma_{i,t}^2}\!\big[\Var(\bphi_t^\top\btheta_{i,t}\mid\sigma_{i,t}^2)\big]\nonumber\\
&\quad+\Var_{\sigma_{i,t}^2}\!\big(\E[\bphi_t^\top\btheta_{i,t}\mid\sigma_{i,t}^2]\big)\nonumber\\
&=\E[\sigma_{i,t}^2]+\E\big[\sigma_{i,t}^2\,\bphi_t^\top\bSigma_t\bphi_t\big]+0\nonumber\\
&=\E[\sigma_{i,t}^2]\,(1+\bphi_t^\top\bSigma_t\bphi_t)\nonumber\\
&=\frac{d_{i,t}}{c_{i,t}-1}(1+\bphi_t^\top\bSigma_t\bphi_t).\label{eq:var_node}
\end{align}
The estimate of the total variance of output $i$ is therefore
\begin{multline}\label{eq:var_mc}
\sigma_{\mathrm{total},i}^2\approx\frac{1}{N_{\mathrm{mc}}}\sum_{m=1}^{N_{\mathrm{mc}}}\big(\hat y_{i,t}^{(m)}-\hat y_{i,t}\big)^2\\
+\Big(1+\frac{1}{N_{\mathrm{mc}}}\sum_{m=1}^{N_{\mathrm{mc}}}\bphi_t^{(m)\top}\bSigma_t\bphi_t^{(m)}\Big)\frac{d_{i,t}}{c_{i,t}-1},
\end{multline}
where $\hat y_{i,t}^{(m)}$ is the $i$th element of $\mathbf{M}_t^\top\bphi_t^{(m)}$.

\section{\texorpdfstring{Proof of Theorem~\ref{thm:regret}}{Proof of Theorem 1}}\label{app:regret}

We consider one output dimension. All $O(\cdot)$ and $o(\cdot)$ statements in this proof hold almost surely.
With the spectral parameters $\{\On,\bL,\bb\}$ fixed and $\gamma=1$, the regret in~\eqref{eq:regret} is given by
\begin{align}
R_T&=-\ln p(y_{1:T}\mid\bx_{1:T})-\inf_{\btheta,\sigma^2}\sum_{t=1}^T\left[-\ln p(y_t\mid\btheta,\sigma^2)\right]\nonumber\\
&=-\ln p(y_{1:T}\mid\bx_{1:T})-\sum_{t=1}^T\left[-\ln p(y_t\mid\hat\btheta_{\mathrm{ML}},\hat\sigma_{\mathrm{ML}}^2)\right],\label{eq:regret_app}
\end{align}
where $\hat\btheta_{\mathrm{ML}}$ and $\hat\sigma_{\mathrm{ML}}^2$ are the maximum likelihood (ML) estimates.

\subsection{The Two Losses}

The marginal likelihood follows from the identity
\begin{equation}\label{eq:marg_id}
p(y_{1:T}\mid\bx_{1:T})=\frac{p(y_{1:T}\mid\bx_{1:T},\btheta,\sigma^2)\,p(\btheta,\sigma^2)}{p(\btheta,\sigma^2\mid y_{1:T},\bx_{1:T})}.
\end{equation}
For the normal inverse-gamma model, we obtain
\begin{equation}\label{eq:marg_lik}
p(y_{1:T}\mid\bx_{1:T})=\frac{1}{(2\pi)^{T/2}}\frac{\sqrt{\det(\bSigma_T)}}{\sqrt{\det(\bSigma_0)}}\frac{\Gamma(c_T)}{\Gamma(c_0)}\frac{d_0^{c_0}}{d_T^{c_T}},
\end{equation}
where $c_T=c_0+T/2$ and $d_T=d_0+\frac{1}{2}\big[\sum_ty_t^2+\mathbf{m}_0^\top\bSigma_0^{-1}\mathbf{m}_0-\mathbf{m}_T^\top\bSigma_T^{-1}\mathbf{m}_T\big]$. The Bayesian loss is therefore
\begin{multline}\label{eq:loss_bayes}
\ell_{\mathrm{Bayes}}(T)=\frac{T}{2}\ln(2\pi)+\frac{1}{2}\ln\frac{\det(\bSigma_0)}{\det(\bSigma_T)}+\ln\frac{\Gamma(c_0)}{\Gamma(c_T)}\\
+c_T\ln d_T-c_0\ln d_0.
\end{multline}
The comparator loss is a Gaussian log loss,
\begin{align}
\sum_{t=1}^T&\left[-\ln p(y_t\mid\btheta,\sigma^2)\right]\nonumber\\
&=\sum_{t=1}^T\left(\frac{1}{2}\ln(2\pi\sigma^2)+\frac{(y_t-\btheta^\top\bphi_t)^2}{2\sigma^2}\right),\label{eq:gauss_loss}
\end{align}
with the ML estimates
\begin{align}
\hat\btheta_{\mathrm{ML}}&=\Big(\sum_{t=1}^T\bphi_t\bphi_t^\top\Big)^{-1}\sum_{t=1}^T\bphi_ty_t,\label{eq:theta_ml}\\
\hat\sigma_{\mathrm{ML}}^2&=\frac{1}{T}\sum_{t=1}^T(y_t-\hat\btheta_{\mathrm{ML}}^\top\bphi_t)^2.\label{eq:sigma_ml}
\end{align}
At these estimates, the comparator loss is
\begin{equation}\label{eq:loss_ml}
\ell_{\mathrm{ML}}=\frac{T}{2}\ln(2\pi)+\frac{T}{2}\ln\hat\sigma_{\mathrm{ML}}^2+\frac{T}{2}.
\end{equation}
We subtract~\eqref{eq:loss_ml} from~\eqref{eq:loss_bayes} and write $c_T\ln d_T=c_0\ln d_T+\frac{T}{2}\ln d_T$. This results in 
\begin{multline}\label{eq:regret_split}
R_T=\frac{1}{2}\ln\frac{\det(\bSigma_0)}{\det(\bSigma_T)}+\ln\frac{\Gamma(c_0)}{\Gamma(c_T)}\\
+c_0\ln\frac{d_T}{d_0}+\frac{T}{2}\ln\frac{d_T}{\hat\sigma_{\mathrm{ML}}^2}-\frac{T}{2}.
\end{multline}
We now examine the terms one at a time.

\subsection{Terms with \texorpdfstring{$d_T$}{d\_T}}

Let $\mathbf{S}_T=\sum_{t=1}^T\bphi_t\bphi_t^\top$. Then
\begin{multline}\label{eq:dT}
d_T=d_0+\frac{T}{2}\hat\sigma_{\mathrm{ML}}^2\\
+\frac{1}{2}(\hat\btheta_{\mathrm{ML}}-\mathbf{m}_0)^\top(\mathbf{S}_T^{-1}+\bSigma_0)^{-1}(\hat\btheta_{\mathrm{ML}}-\mathbf{m}_0).
\end{multline}
The choice of $\mathbf{m}_0$ and $\bSigma_0$ affects only the $O(1)$ terms, so we set $\mathbf{m}_0=\mathbf{0}$ and $\bSigma_0=\bI$ and write
\begin{align}
d_T&=\frac{T}{2}\hat\sigma_{\mathrm{ML}}^2+A_T,\nonumber\\
A_T&=d_0+\frac{1}{2}\hat\btheta_{\mathrm{ML}}^\top(\mathbf{S}_T^{-1}+\bI)^{-1}\hat\btheta_{\mathrm{ML}}.\label{eq:dT2}
\end{align}
By the ergodic theorem and Assumption~\ref{ass:ergodic}, almost surely,
\begin{equation}\label{eq:lln}
\begin{gathered}
\frac{1}{T}\sum_{t=1}^T\bphi_t\bphi_t^\top\to\mathbf{R}_{\phi\phi},\qquad
\frac{1}{T}\sum_{t=1}^T\bphi_t y_t\to\mathbf{r}_{\phi y},\\
\frac{1}{T}\sum_{t=1}^T y_t^2\to\E[y_t^2],
\end{gathered}
\end{equation}
where $\mathbf{r}_{\phi y}=\E[\bphi_t y_t]$. Hence
$\hat{\btheta}_{\mathrm{ML}}\to\mathbf{R}_{\phi\phi}^{-1}\mathbf{r}_{\phi y}$ and
$\hat\sigma^2_{\mathrm{ML}}\to\sigma_*^2>0$ almost surely, so
$\hat\btheta_{\mathrm{ML}}=O(1)$, $\mathbf{S}_T=O(T)$, and
$\mathbf{S}_T^{-1}=O(T^{-1})$ in the Frobenius norm.

With $(\bI+\mathbf{E})^{-1}=\bI-\mathbf{E}+O(\|\mathbf{E}\|^2)$,
\begin{align}
A_T&=d_0+\frac{1}{2}\|\hat\btheta_{\mathrm{ML}}\|_2^2-\frac{1}{2}\hat\btheta_{\mathrm{ML}}^\top\mathbf{S}_T^{-1}\hat\btheta_{\mathrm{ML}}+O(T^{-2})\nonumber\\
&=O(1).\label{eq:AT}
\end{align}
We write $d_T=\frac{T}{2}\hat\sigma_{\mathrm{ML}}^2\big(1+\frac{2A_T}{T\hat\sigma_{\mathrm{ML}}^2}\big)$ and expand the logarithm,
\begin{equation}\label{eq:lndT}
\ln d_T=\ln\frac{T}{2}+\ln\hat\sigma_{\mathrm{ML}}^2+\frac{2A_T}{T\hat\sigma_{\mathrm{ML}}^2}+O(T^{-2}).
\end{equation}
Because $\hat\sigma_{\mathrm{ML}}^2$ converges to a positive limit and $A_T=O(1)$, this expansion gives the two terms with $d_T$ in~\eqref{eq:regret_split},
\begin{align}
c_0\ln\frac{d_T}{d_0}&=c_0\ln T+O(1),\label{eq:dT_term1}\\
\frac{T}{2}\ln\frac{d_T}{\hat\sigma_{\mathrm{ML}}^2}-\frac{T}{2}&=\frac{T}{2}\ln\frac{T}{2}-\frac{T}{2}+\frac{A_T}{\hat\sigma_{\mathrm{ML}}^2}+O(T^{-1})\nonumber\\
&=\frac{T}{2}\ln\frac{T}{2}-\frac{T}{2}+O(1).\label{eq:dT_term2}
\end{align}

\subsection{Terms with \texorpdfstring{$c_T$}{c\_T}}

For $z>0$, the Stirling expansion takes the form
\begin{equation}\label{eq:stirling}
\ln\Gamma(z)=\Big(z-\frac{1}{2}\Big)\ln z-z+\frac{1}{2}\ln(2\pi)+\mu(z),
\end{equation}
where $\mu(z)\to0$ as $z\to\infty$. With $z=c_T=c_0+T/2$ and $\ln c_T=\ln\frac{T}{2}+\frac{2c_0}{T}+O(T^{-2})$, we obtain
\begin{align}
\ln\Gamma(c_T)&=\Big(c_0+\frac{T}{2}-\frac{1}{2}\Big)\Big(\ln\frac{T}{2}+\frac{2c_0}{T}\Big)\nonumber\\
&\quad-c_0-\frac{T}{2}+O(1)\nonumber\\
&=\frac{T}{2}\ln\frac{T}{2}-\frac{T}{2}+\Big(c_0-\frac{1}{2}\Big)\ln T+O(1).\label{eq:lngamma}
\end{align}
Hence
\begin{multline}\label{eq:gamma_term}
\ln\frac{\Gamma(c_0)}{\Gamma(c_T)}=-\frac{T}{2}\ln\frac{T}{2}+\frac{T}{2}\\
-\Big(c_0-\frac{1}{2}\Big)\ln T+O(1).
\end{multline}

When we add~\eqref{eq:dT_term1}, \eqref{eq:dT_term2}, and~\eqref{eq:gamma_term}, the terms $\pm\frac{T}{2}\ln\frac{T}{2}$, $\pm\frac{T}{2}$, and $\pm c_0\ln T$ cancel, and we are left with
\begin{multline}\label{eq:noise_part}
\ln\frac{\Gamma(c_0)}{\Gamma(c_T)}+c_0\ln\frac{d_T}{d_0}+\frac{T}{2}\ln\frac{d_T}{\hat\sigma_{\mathrm{ML}}^2}-\frac{T}{2}\\
=\frac{1}{2}\ln T+O(1).
\end{multline}
This term is the cost of the unknown noise variance.

\subsection{\texorpdfstring{Term with $\bSigma_T$}{Term with Sigma\_T}}

The recursion $\bSigma_{t+1}=\bSigma_t-\mathbf{k}_t(\bSigma_t\bphi_t)^\top$ in Algorithm~\ref{alg:rbr} is equivalent to
\begin{equation}\label{eq:info}
\bSigma_T^{-1}=\bSigma_0^{-1}+\mathbf{S}_T=T\Big(\frac{1}{T}\bSigma_0^{-1}+\bar{\mathbf{S}}_T\Big),
\end{equation}
where $\bar{\mathbf{S}}_T=\mathbf{S}_T/T$. Hence $\det(\bSigma_T^{-1})=T^J\det\big(\bar{\mathbf{S}}_T+\frac{1}{T}\bSigma_0^{-1}\big)$, and with $\bSigma_0=\bI$,
\begin{align}
\frac{1}{2}\ln\frac{\det(\bSigma_0)}{\det(\bSigma_T)}&=\frac{J}{2}\ln T+\frac{1}{2}\ln\det\Big(\bar{\mathbf{S}}_T+\frac{1}{T}\bI\Big)\nonumber\\
&=\frac{J}{2}\ln T+\frac{1}{2}\ln\det(\mathbf{R}_{\phi\phi})+o(1),\label{eq:sigma_term}
\end{align}
where the last step uses~\eqref{eq:lln} and the positive definiteness of $\mathbf{R}_{\phi\phi}$ in Assumption~\ref{ass:ergodic}. This term is the cost of the $J$ unknown weights.

\subsection{Conclusion of the Proof}

We combine~\eqref{eq:noise_part} and~\eqref{eq:sigma_term} in~\eqref{eq:regret_split} and obtain
\begin{equation}\label{eq:regret_final}
R_T=\frac{J}{2}\ln T+\frac{1}{2}\ln T+O(1)=\frac{J+1}{2}\ln T+O(1),
\end{equation}
which completes the proof. \IEEEQED

\ifinternal
\section{Gradients (Internal Check, Not for Submission)}\label{app:grad}

Because our model is Bayesian, the data term is the negative log-likelihood of a Student's $t$ distribution,
\begin{multline*}
\mathcal{L}_{\mathrm{NLL},i}=\frac{\nu+1}{2}\ln\!\left(1+\frac{\tilde y_{i,t}^2}{\nu\sigma_{i,t}^2}\right)+\ln\Gamma\!\left(\frac{\nu}{2}\right)\\
-\ln\Gamma\!\left(\frac{\nu+1}{2}\right)+\frac{1}{2}\ln(\pi\nu\sigma_{i,t}^2),
\end{multline*}
where $\nu=2c_{i,t}$ is the number of degrees of freedom, $\sigma_{i,t}^2=s_td_{i,t}/c_{i,t}$ is the scale, and $\tilde y_{i,t}$ is the residual. In batch learning, the coefficients of the structural penalties in~\eqref{eq:loss} are proportional to the number of training samples. For online learning, we introduce explicit weights $\beta_m$ and $\beta_s$,
\begin{multline}\label{eq:loss_online}
\ell_{\mathrm{total}}=-\E_q[\ln p(\by\mid\On,\bL,\bb)]+\beta_m\|\Mb^\top\Mb-\bI\|_F^2\\
+\beta_s\sum_{j,d}\left(\lambda S_{dj}^2-\ln S_{dj}\right).
\end{multline}
From a Bayesian point of view, $\beta_m$ and $\beta_s$ act as inverse temperatures of the H-DPP prior. The weight $\beta_m$ sets the effective observation horizon; for instance, $\beta_m=1/1000$ matches the influence of the prior to the most recent 1{,}000 samples. For consistency with the batch objective, we set $\beta_s=2\beta_m$, but the two weights can also be tuned separately to balance orthogonality and variance regularization.

We update $\{\On,\bL,\bb\}$ by SGD and propagate the error of $\mathcal{L}_{\mathrm{NLL}}$ back through the feature map as follows.

\emph{Output.} The gradient with respect to the predictive mean $\hat y_{i,t}$ is
\begin{equation}
g_{y,i}=\frac{\partial\mathcal{L}_{\mathrm{NLL},i}}{\partial\hat y_{i,t}}=-\frac{(\nu+1)(y_{i,t}-\hat y_{i,t})}{\nu\sigma_{i,t}^2+(y_{i,t}-\hat y_{i,t})^2}.
\end{equation}
This gradient automatically down-weights outliers with large squared residuals.

\emph{Features.} Through the mean matrix $\mathbf{M}_t$, $\mathbf{g}_\phi=\mathbf{M}_t\mathbf{g}_y\in\R^J$.

\emph{Activations.} With $\bphi_t=\sqrt{2/J}\cos(\mathbf{z}_t)$, $\mathbf{g}_z=-\sqrt{2/J}\,\mathbf{g}_\phi\odot\sin(\mathbf{z}_t)\in\R^J$.

\emph{Frequencies.} $\mathbf{G}_\Omega=\bx_t\mathbf{g}_z^\top\in\R^{\Dx\times J}$.

\emph{Lengthscales.} With $\bL=\diag(\bm{\ell})$ and $\bOmega=\bL^{-1}\On$,
\begin{equation}
g_{\ell_d}=-\frac{1}{\ell_d}\sum_{j=1}^J[\mathbf{G}_\Omega]_{dj}\Omega_{dj}.
\end{equation}
To keep the lengthscales positive, we update them in the log domain with $\mathbf{g}_{\ln\ell}=-\mathrm{rowsum}(\mathbf{G}_\Omega\odot\bOmega)$.

\emph{Phases.} $\mathbf{g}_b=\mathbf{g}_z$.

\emph{Standardized frequencies.} $\mathbf{G}_{\On}=\diag(1/\bm{\ell})\,\mathbf{G}_\Omega$.

\emph{Projected mean.} The gradient with respect to $\Mb$ combines the data term and the repulsion,
\begin{equation}
\mathbf{G}_{\Mb}=\mathbf{G}_{\On}+4\beta_m\Mb(\Mb^\top\Mb-\bI).
\end{equation}

\emph{Unconstrained mean.} For each column, the chain rule through $\mbar_j=\mathbf{m}_j/\|\mathbf{m}_j\|$ gives
\begin{equation}
\mathbf{g}_{\mathbf{m}_j}=\frac{1}{\|\mathbf{m}_j\|}\left(\mathbf{g}_{\mbar_j}-(\mbar_j^\top\mathbf{g}_{\mbar_j})\mbar_j\right).
\end{equation}
Because we renormalize after every step, $\|\mathbf{m}_j\|=1$ at the next step, and the gradient reduces to the projection of $\mathbf{g}_{\mbar_j}$ onto the tangent space of the sphere. The update therefore moves along the surface of the sphere and limits numerical drift.

\emph{Standard deviations.} The data term gives $\mathbf{G}^{\mathrm{NLL}}_{\Sn}=\mathbf{G}_{\On}\odot\En$, and the regularizer gives $\mathbf{G}^{\mathrm{KL}}_{\Sn}=2\beta_s\lambda\Sn-\beta_s/\Sn$. In the log domain, which keeps $S_{dj}>0$,
\begin{equation}
\mathbf{G}_{\ln\Sn}=(\mathbf{G}_{\On}\odot\En+2\beta_s\lambda\Sn)\odot\Sn-\beta_s.
\end{equation}
The prediction error drives $\Sn$, and the prior prevents both its collapse to a point estimate and its unbounded growth.

\emph{Learning rate (hypergradient descent).} If the current gradient points in the same direction as the previous one, we accelerate; if it points in the opposite direction, the step is too large or the data have changed, and we decelerate. For the SGD update $\btheta_t=\btheta_{t-1}-\eta_{t-1}\nabla\ell_{t-1}(\btheta_{t-1})$, the chain rule gives
\begin{equation}
\frac{\partial\ell_t}{\partial\eta_{t-1}}=-\nabla\ell_t\cdot\nabla\ell_{t-1}.
\end{equation}
A gradient step on $u=\ln\eta$ with meta step size $\alpha$ yields
\begin{equation}
\ln\eta_t=\ln\eta_{t-1}+\alpha\,(\nabla\ell_t\cdot\nabla\ell_{t-1}).
\end{equation}
\fi

\end{document}